\documentclass[journal]{IEEEtran}

\usepackage{times}  
\usepackage{helvet}  
\usepackage{courier}  
\usepackage{url}  
\usepackage{graphicx}  
\usepackage{bm}
\usepackage{amsmath,amssymb,amsfonts}
\usepackage{multirow}
\usepackage{multicol}
\usepackage{enumitem}

\ifCLASSINFOpdf
\else
\fi
\begin{document}

\newcommand{\drr}{\emph{DRr-Net}}
\newcommand{\fdrr}{\emph{Dynamic Re-read Network (DRr-Net)}}

\newcommand{\dlan}{\emph{LadRa-Net}}
\newcommand{\fdlan}{\emph{Locally-Aware Dynamic Re-read Attention Net~(LadRa-Net)}}
%
\title{LadRa-Net: Locally-Aware Dynamic Re-read Attention Net for Sentence Semantic Matching}
%
%
%

\author{Kun~Zhang~\IEEEmembership{Member,~IEEE}, Guangyi~Lv, Le~Wu~\IEEEmembership{Member,~IEEE}, 
	Enhong~Chen,~\IEEEmembership{Senior~Member,~IEEE},
	Qi~Liu~\IEEEmembership{Member,~IEEE},
	and Meng Wang,~\IEEEmembership{Fellow,~IEEE}
	\IEEEcompsocitemizethanks{\IEEEcompsocthanksitem K. Zhang, L Wu, and M Wang are with Key Laboratory of Knowledge Engineering with Big Data, and School of Computer and Information, Hefei University of Technology, Hefei, Anhui 230029, China. (email: zhang1028kun, lewu.ustc, eric.mengwang@gmail.com).
	\IEEEcompsocthanksitem G Lv is with AI Lab at Lenovo Research, Beijing, 100094, China.
	\IEEEcompsocthanksitem G Lv, E Chen, and Q Liu are with University of Science and Technology of China, Hefei 230026, China. (email: gylv@mail.ustc.edu.cn, cheneh, qiliuql@ustc.edu.cn).}
}
%
%

\markboth{Journal of \LaTeX\ Class Files,~Vol.~14, No.~8, August~2015}%
{Shell \MakeLowercase{\textit{et al.}}: Bare Demo of IEEEtran.cls for IEEE Journals}
%



\maketitle

 \begin{abstract}
	Sentence semantic matching requires an agent to determine the semantic relation between two sentences, which is widely used in various natural language tasks, such as Natural Language Inference~(NLI), Paraphrase Identification~(PI), and so on. 
	Much recent progress has been made in this area, especially attention-based methods and pre-trained language model based methods.    
	However, most of these methods focus on all the important parts in sentences in a static way and only emphasize how important the words are to the query, inhibiting the ability of attention mechanism.
	In order to overcome this problem and boost the performance of attention mechanism, we propose a novel dynamic re-read attention, which can pay close attention to one small region of sentences at each step and re-read the important parts for better sentence representations. 
	Based on this attention variation, we develop a novel \fdrr~for sentence semantic matching.
	Moreover, selecting one small region in dynamic re-read attention seems insufficient for sentence semantics, and employing pre-trained language models as input encoders will introduce incomplete and fragile representation problems. 
	To this end, we extend \drr~to \fdlan, in which local structure of sentences is employed to alleviate the shortcoming of Byte-Pair Encoding~(BPE) in pre-trained language models and boost the performance of dynamic re-read attention. 
	Extensive experiments on two popular sentence semantic matching tasks demonstrate that \drr~can significantly improve the performance of sentence semantic matching. 
	Meanwhile, \dlan~is able to achieve better performance by considering the local structures of sentences. 
	In addition, it is exceedingly interesting that some discoveries in our experiments are consistent with some findings of psychological research.
\end{abstract}

\begin{IEEEkeywords}
Sentence Semantic Matching, Dynamic Re-read Attention, Local Strucute, Representation Learning.
\end{IEEEkeywords}

%
\IEEEpeerreviewmaketitle

\section{Introduction}
\label{s:introduction}
\begin{figure}
	\centering
	\includegraphics[width=0.45\textwidth]{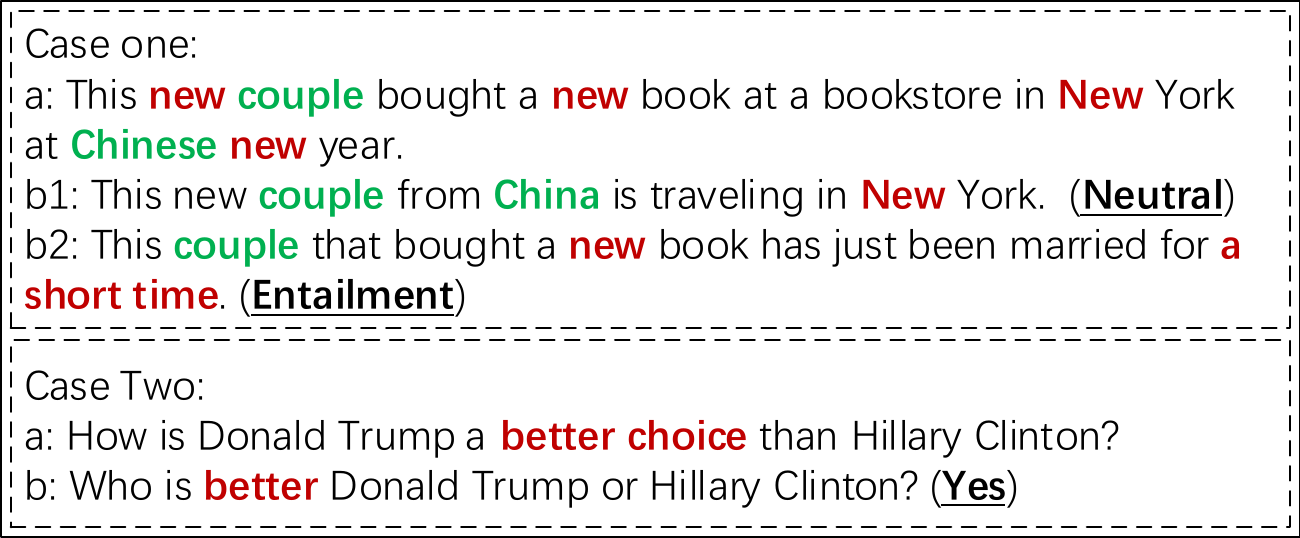}
	\caption{The examples for SNLI and Quora datasets.} 
	\label{f:example}
	\vspace{-4mm}
\end{figure}

\IEEEPARstart{S}{entence} semantic matching is a fundamental technology in Natural Language Processing, which requires an agent to predict the semantic relation between two sentences. 
For example, in Natural Language Inference~(NLI), sentence semantic matching is leveraged to determine whether a hypothesis sentence can be inferred reasonably from a given premise sentence~\cite{Kim2018SemanticSM}. 
In Paraphrase Identification~(PI), it is utilized to identify whether two sentences have identical meaning or not~\cite{dolan2005automatically}. 
Fig.~\ref{f:example} illustrates two representative examples about NLI and PI, respectively. 

As a fundamental yet challenging task, sentence semantic matching has been applied in many fields, e.g., information retrieval~\cite{Clark2016CombiningRS}, question answering~\cite{wang2017vqa,liu2018finding}, and dialog system~\cite{serban2016building}. 
With the large annotated datasets, such as SNLI~\cite{bowman2015large}, SCITAIL~\cite{khot2018scitail}, Quora~\cite{iyer2017first}, and MSRP~\cite{dolan2005automatically}, and advancement of representation learning techniques, such as CNN~\cite{kim2014convolutional}, LSTM~\cite{Cheng2016LongSM}, GRU~\cite{Chung2014EmpiricalEO}, and Attention Mechanism~\cite{vaswani2017attention,zhang2018ImageEnhance}, rapid development on sentence semantic matching has been enabled. 
Early work often focuses on designing different structures for matching modeling~\cite{Kim2018SemanticSM,zhang2018ImageEnhance,talman2018natural}.
Recently, pre-trained language models~(BERT~\cite{devlin2018bert}, GPT~\cite{Radford2018ImprovingLU}) have become the new scheme for language understanding. Researchers employ these pre-trained methods to process the input sentences and design task-aware network structures for the final tasks, which have achieved much progress in many NLP tasks.

However, there are still some limitations in most existing methods. 
First of all, most of them adopt attention mechanism to select all the important parts of sentences in a static way, restricting the ability of attention mechanism. 
Inspired by the idea that pre-trained methods leverage dynamic embedding methods to replace the static embedding methods~(e.g., Word2Vec~\cite{Mikolov2013DistributedRO}, Glove~\cite{Pennington2014GloveGV}), we intend to make full use of attention mechanism dynamically, which is one of our main contributions. 
For this direction, human reading behaviors can bring us plenty of inspirations. 
``\textit{When reading a piece of text, people will constantly change their focal point for in-depth understanding based on what they already know}.''
Taking case one in Fig.~\ref{f:example} as an example, when classifying the inference relation between \textit{a} and \textit{b1}, we first try to understand the global meaning of the sentence in a sequential manner (i.e., reading it word by word) and grasp information as much as possible, such as the object: \textit{new couple} and \textit{new book}, the place: \textit{bookstore} and \textit{New York}, and the time: \textit{Chinese new year}. 
After reading~\textit{b1}, we will pay more attention to the object and the place. 
When confirming that the place is the same, we will turn to identify whether the couple comes from China. 
Finally, we can conclude that the relation is \textit{Neutral}. 
Meanwhile, psychological research has also shown that there are plenty of reading behaviors of humans for efficient language understanding and communication. 
For example, humans only pay attention to a small region of information at one time, i.e., people only focus on 1.5 words each time when intensively reading a piece of text~\cite{wang1999reading}. 
Koch et al.~\cite{koch2007attention} has demonstrated that people focused on less than 7 different objects at the same time. 
Therefore, how to capture these important parts in a dynamic way and make full use of them for in-depth semantic understanding is one of the challenges that we should focus on.

Another easily overlooked limitation is the usage of Byte-Pair Encoding (BPE) or its variations in BERT based methods. 
Though this encoding method is capable of avoiding OOV~(Out-Of-Vocab) problem, it splits a word into subword units, which suffers from two problems: \textit{incomplete modeling} and \textit{fragile representation}~\cite{Ma2020CharBERTCP}. 
Therefore, we try to leverage local structures of input sentences to alleviate these problems and further improve the ability of our proposed dynamic attention mechanism. 
For this direction, we can also learn from human reading behaviors. 
As shown in Fig.~\ref{f:example}, when predicting the inference relation between \textit{a} and \textit{b2}, we will focus on the \textit{couple} and try to figure out how long they have been married. 
Traditional methods may be confused by the four ``\textit{new}''s in \textit{a}. 
When taking the local structure into consideration, we obtain that the first ``\textit{new}'' is the attribute of the \textit{couple}, indicating they are newly-married.
Thus, the inference relation between \textit{a} and \textit{b2} is ``\textit{Entailment}''. 
Biological researchers have also observed that people will focus on what they care about most, and leverage peripheral vision~\cite{brandt1973differential,warren1992role} to get the local structure for understanding enhancement. 
Similar observations have been found by Zheng et al.~\cite{zheng2019human} through a lab study.  
To this end, how to select and use the local structure effectively while paying close attention to important words is another problem that we need to tackle.

In order to solve the challenge of how to employ attention mechanism in a dynamic way for enhancing its ability of selecting important parts, in our preliminary work~\cite{zhang2019drr}, we proposed the \textit{Dyanmic Re-read~(DRr)} attention, a novel architecture that selects one important word at each reading step and reads the important words repeatedly for precise sentence semantic representation. 
Based on this novel attention, we developed a \fdrr~for sentence semantic matching, in which Global Sentence Encoding is used to model sentences comprehensively and \textit{DRr} unit is used to capture the important parts precisely for better sentence semantic representation and matching.
In this way, \drr~was able to select the most important word to process based on the learned information, which was in favor of tackling sentence semantic matching task.

However, there still exists some space in \drr~for further improvement. 
First, the encoding ability of Global Sentence encoding is insufficient for input embedding. 
Replacing with pre-trained methods~(e.g., BERT) will cause the problem that BPE suffers from on the other hand. 
Second, selecting only one important word at each reading step in \drr~may be insufficient since words can express tremendously different meanings with different local structures.
Therefore, in this paper, we focus on using local structure to alleviate the above problems. 
We propose the \fdlan, an advanced architecture that leverages pre-trained language models to encode input sentences and uses the local structure to alleviate the weaknesses of BPE in pre-trained methods and further enhance the ability of our proposed Dynamic Re-read attention. 
In concrete details, we first take advantage of pre-trained BERT~\cite{devlin2018bert} to encode input sentences from a global perspective. 
Meanwhile, we develop a \emph{Phrase-CNN} unit to capture local structures of the input sentence with different composite kernels~(e.g., bigram and trigram) over the output of BERT, so that the input sentences can be encoded sufficiently. 
After getting the words and local phrase representations, 
a newly designed \emph{Dynamic Sequential Attention~(DSA)} unit is employed to perform dynamic attention in a sequential manner. 
Along this line, not only the important parts can be measured dynamically, but also the corresponding local structure can be fully explored. 
Meanwhile, the ability of BPE in pre-trained models can also be strengthened.
Therefore, \dlan~is capable of making better modeling of sentence semantics and doing better prediction of sentence semantic relations.

The remainder of this paper is organized as follows. 
We will first introduce the related work in Sections~\ref{s:related-work}. 
Then, in Section~\ref{s:problem}, we give a formal definition of sentence semantic matching task and propose two important issues that will be tackled in this paper. 
Next, the structure and technical details of our proposed models are given in Sections~\ref{s:drr} and~\ref{s:dlan}.  
The experiments and detailed analysis are given in Section~\ref{s:experiment}. 
Finally, we discuss and conclude our work in Section~\ref{s:conclusion}.

\section{Related Work}
\label{s:related-work}

Our work is related to two lines of literature, 1) \textit{Sentence Semantic Matching}: focusing on the semantic representations and relation verification of sentence pairs; 2) \textit{Human Attention Behavior}: focusing on the study of attention mechanism in human reading behavior.

\subsection{Sentence Semantic Matching}
Based on the learning schema, this part can be grouped into two categories: \textit{traditional neural network based methods} and \textit{pre-trained language model based methods}. We will introduce each of them in the following parts:

\subsubsection{\textbf{Traditional Neural Network based methods}}
With the developments of various neural network technologies such as CNN~\cite{kim2014convolutional}, RNN~\cite{Cheng2016LongSM}, attention mechanism~\cite{vaswani2017attention,wu2019hierarchical} as well as Graph Neural Network~(GNN)~\cite{song2020kganet,wu2020joint}, plenty of methods have been exploited to model semantic matching on large datasets like SNLI~\cite{bowman2015large}, SciTail~\cite{khot2018scitail}, and Quora~\cite{iyer2017first}.
Traditionally, researchers try to make full use of neural network technologies to model sentence semantic meanings in an end-to-end fashion. 
Among them, RNNs mainly focus on the sequential information and the semantic dependency of sequences~\cite{Liu2016LearningNL,peng2020enhanced,hu2020enhanced}.
CNNs tend to capture the local context with different convolutional filters~\cite{kim2014convolutional,xu2020enhanced}. 
Attention mechanism is usually utilized to extract the most important parts in sentences, capture the semantic relations and align the elements of two sentences properly~\cite{Cho2015DescribingMC,zhang2017context}.
For example, Tay et al.~\cite{tay2018co} employed a stacked multi-layer Bi-LSTM with Alignment Factorization to measure all feature hierarchies among two sentences.
Yang et al.~\cite{yang2019simple} utilized multi-layer encoding and fusion block based on CNN structure to build a fast and well-performed sentence matching model.
KIM et al.~\cite{Kim2018SemanticSM} proposed a co-attention network to model the interaction between two sentences, and developed a densely-connected structure to retain as much information as possible.
Dong et al.~\cite{dong2020distilling} adopted GNN to access the structure information of input sentences for comprehensive sentence relation modeling. 
Kun et al.~\cite{zhang2018ImageEnhance} combined these three architectures into a hybrid architecture, in which CNN is utilized to generate phrase-level semantic meanings, GRU is used for the word sequence and dependency among sentences, and attention is adopted to integrate different features for the final classification.

\subsubsection{\textbf{Pre-trained Language Model based methods}}
In order to make full use of existing large language corpora, various pre-trained language models have been proposed.
For example, Vaswani et al.~\cite{vaswani2017attention} proposed the transformer architecture, in which multi-head self-attention and residual connection are used to analyze the inputs. 
By stacking multiple layers of transformers, input sentences could be fully explored. 
Based on the transformer and large language corpora, GPT~\cite{Radford2018ImprovingLU} and BERT~\cite{devlin2018bert} were proposed to learn language representations in a pre-trained manner. 
Among them, BERT achieved a better performance. 
It made full use of self-supervised Mask Language Model~(MLM) and Next Sentence Prediction~(NSP) tasks to get a comprehensive understanding of input sentences. 
Moreover, by modifying the downstream tasks to the required form, BERT was capable of providing promising representations for inputs and improving the model performance effectively~\cite{devlin2018bert}. 
To further improve the performance of pre-trained models, one direction is modifying the input encoding and self-supervised pre-trained task usage. 
For example, XLNet~\cite{Yang2019XLNetGA} leveraged a newly designed Permutation Language Model~(PLM) task to narrow down the gap between pre-trained tasks and downstream tasks. 
In addition, there still exists much promising work, such as RoBERTa~\cite{Liu2019RoBERTaAR}, CharBERT~\cite{Ma2020CharBERTCP}, as well as BART~\cite{lewis2020bart}. 
Moreover, pre-trained methods still have some weaknesses in accessing external knowledge. 
To this end, many external knowledge enriched methods have been proposed. such as incorporating knowledge base~\cite{Zhang2019ERNIEEL}, integrating entity knowledge into pre-trained stage~\cite{Sun2019ERNIEER}, as well as considering syntax and semantics knowledge~\cite{liu2020sentence}.

The above work has made great progress in sentence semantic matching and inspired us to fully utilize advanced neural networks and pre-trained methods for  sentence semantic modeling. 
On the other hand, current methods still have some weaknesses in dealing with sentence semantics and need to be further improved. 
First, Wen et al.~\cite{Ma2020CharBERTCP} has proved that BPE structure in BERT is insufficient for word representations and easy to be attacked. 
Meanwhile, attention operation in BERT is calculated among the entire input sequence, in which the irrelevant parts would interfere the weights of important parts. 
Thus, better modeling for input embedding should be considered. 
Second, most existing methods only treat the structured information~(Knowledge Base) as prior knowledge, ignoring the potential of reading behaviors of humans. 
Humans invent many reading behaviors~(e.g., skipping and repeating) for language understanding and communication efficiently~\cite{zheng2019human}, which can also be treated as prior knowledge. 
Therefore, in this paper, we take human reading behaviors into consideration and 
propose the novel \textit{DRr} and \textit{DSA} units to boost the ability of attention mechanism for better sentence semantic modeling. 
And we design two sentence semantic matching models~(i.e., \drr~and \dlan) based on \textit{DRr} and \textit{DSA} units separately to tackle the above problems.

\subsection{Human Attention Behavior}

Human learning has inspired various algorithm designs throughout the development of machine learning~\cite{wang2020comprehensive}. For example, Curriculum Learning~\cite{xu2020curriculum} tries to train a model from easier data to harder data, which imitates the meaningful learning order in human curricula. 
Attention mechanism helps an agent to focus on the most relevant parts of input for output to achieve better performance, which imitates the human ability of quickly perceiving necessary information by selectively attending to parts of what they saw~\cite{das2016human}.
Despite the success of imitating human actions, we have to admit that these algorithm designs still have large performance gaps with human beings in more practical settings~\cite{zheng2019human}. Learning better from human behaviors may help models to achieve better performances. 

Taking attention mechanism as an example, Zheng et al.~\cite{zheng2019human} observed some user behavior patterns for better attention usage with a lab study. 
For example, in a specific scenario (e.g., Answer Selection), users tended to pay more attention to possible segments that are relevant to what they want. They would re-read more snippets of candidate answers with more \textit{skip} and \textit{up} behaviors, while ignoring the irrelevant parts~\cite{zheng2019human,li2019teach}. 
Moreover, Sen et al.~\cite{sen2020human} obtained significant similarities between human attention and machine attention on a large text classification dataset by experiments. 
Peng et al.~\cite{peng2020bi} demonstrated that imitating reverse thinking and inertial thinking of humans can improve the model performance on reading comprehension task. 
These observations indicated that human attention was helpful for guiding neural network design and it is a very promising direction.

Besides, psychologists also have similar observations. 
For example, Yarbus~\cite{yarbus1967eye} described that human attention ``\textit{is dependent on not only what is shown in the picture, but also the problem facing the observer and the information that he hopes to gain}''. 
By building an eye tracker, Wang et al.~\cite{wang1999reading} found that people may center on 1.5 words each time when reading a piece of text intensively. 
In addition, researchers had already demonstrated that human eyes have central vision and peripheral vision. 
Central vision concentrated on what a person needs at the current time. 
Peripheral vision used the coarse-grained observation of the surroundings to support the central vision~\cite{brandt1973differential,warren1992role,strasburger2011peripheral}. 

These studies encourage us to learn from human learning behaviors and improve the model performance by incorporating this behavior knowledge. 
Therefore, in this paper, we focus on utilizing behavior knowledge to further improve the model performance on sentence semantic modeling.

\section{Problem Statement}
\label{s:problem}
In this section, we formulate the sentence semantic matching task as a supervised classification problem. 
Given the triple $(\bm{s}^a, \bm{s}^b, y)$, where $\bm{s}^a = \{\bm{w}_1^a, \bm{w}_2^a, ..., \bm{w}_{l_a}^a \}$ and $\bm{s}^b = \{\bm{w}_1^b, \bm{w}_2^b, ..., \bm{w}_{l_b}^b \}$ are the given sentences, $\bm{w}_i^a$ and $\bm{w}_j^b$ are one-hot vectors which represent the $i^{th}$ and $j^{th}$ word in the sentences, and $l_a$ and $l_b$ indicate the total number of words in $\bm{s}^a$ and $\bm{s}^b$. 
The true label $y\in\mathcal{Y}$ indicates the semantic relationship between the given sentence pair, where $\mathcal{Y}=\{entailment, contradiction, neutral\}$ for NLI task and $\mathcal{Y}=\{Yes, No\}$ for PI task. 
Our goal is to learn the classifier $\xi$ which is able to compute the conditional probability $P(y|\bm{s}^a, \bm{s}^b)$ and predict the label for the given sentence pair in the test set by $y^{*}=argmax_{y\in\mathcal{Y}}P(y|\bm{s}^a, \bm{s}^b)$.

\begin{figure*}
	\centering
	\includegraphics[width=0.75\textwidth]{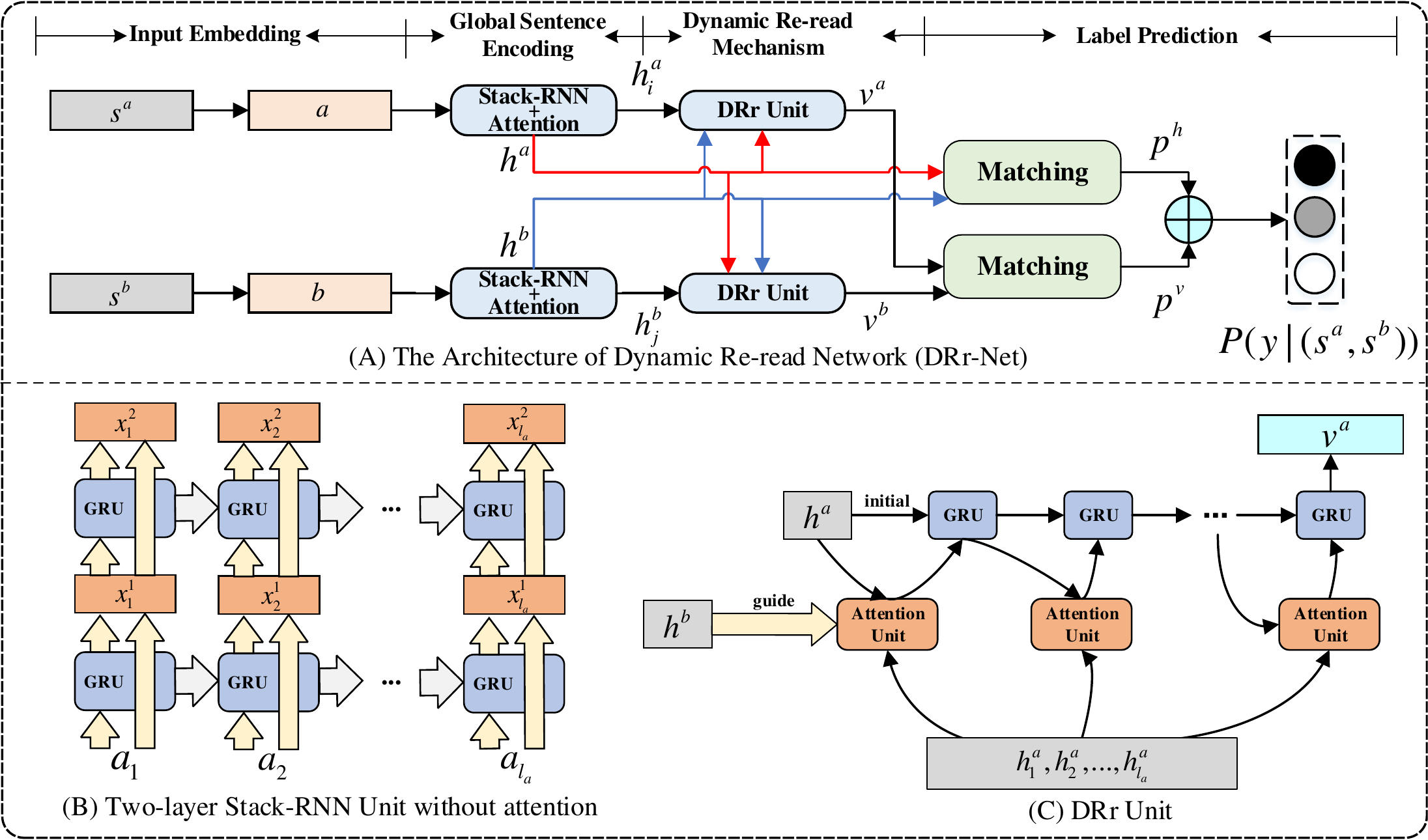}
	\caption{(A): Architecture of \fdrr. { (B): Stack-RNN processes the original sentence and preserves all the information from bottom-most word embedding input to up-most recurrent output. (C): DRr Unit pays close attention to only one important word at each step with the learned information and re-read these important words for sentence semantic matching more precisely.}}
	\label{f:drr_model}
	\vspace{-4mm}
\end{figure*}

In order to do better sentence semantic matching, the following important challenges should be considered:
\begin{itemize}\setlength{\itemsep}{0pt}
	\item Static usage of attention may inhibit the ability of attention mechanism. How to adopt attention mechanism dynamically for important parts selection and sentence semantic understanding?
	
	\item The local structures are helpful for overcoming the weaknesses of BPE in BERT and enhancing the ability of DRr attention. How to select and use this information efficiently while selecting the important words?
	
\end{itemize}

To this end, we propose \drr~and \dlan~to tackle the above issues for sentence semantic matching.

\section{\fdrr}
\label{s:drr}

The overall architecture is shown in Fig.~\ref{f:drr_model}(A), which contains four main components: 1) \textit{Input Embedding}, 2) \textit{Global Sentence Encoding}, 3) \textit{Dynamic Re-read}, and 4) \textit{Label Prediction}. Next, we will introduce each of them in details.

\subsection{Input Embedding}
The Input Embedding converts each word of sentences into a vector representation and constructs the representation matrix for the sentences. 
We combine multiple features as the semantics representations of words. 
To be specific, the inputs of \drr~are one-hot representations $\bm{s}^a = \{\bm{w}_1^a, \bm{w}_2^a, ..., \bm{w}_{l_a}^a \}$ and $\bm{s}^b = \{\bm{w}_1^b, \bm{w}_2^b, ..., \bm{w}_{l_b}^b \}$. 
For more comprehensive access to the semantics of each word in sentences, we adopt the concatenation of pre-trained word embedding~\cite{pennington2014glove}, character features~\cite{gong2017natural}, and syntactical features~\cite{chen2017reading,gururangan2018annotation} to represent each word in sentences. 
The word embedding is obtained by mapping each token to high dimensional vector space by pre-trained word vector (840B Glove~\cite{pennington2014glove}). 
The character features are obtained by applying a CNN with a max-pooling layer to the learned character embeddings, which can represent words in a finer-granularity and help to avoid the Out-Of-Vocabulary problem that pre-trained word vectors often suffer from. 
The syntactical features consist of one-hot Part-Of-Speech~(POS) tagging feature, binary exact match feature, and binary antonym feature, which have been proved useful for sentence semantic understanding~\cite{chen2017reading}. 
Next, we pass these representations through a two-layer highway network~\cite{srivastava2015highway} and get the extravagant representations $\{\bm{a}_i|i = 1, 2, ..., l_a\}$ and $\{\bm{b}_j|j=1, 2, ..., l_b\}$ for the words in sentences $\bm{s}^a$ and $\bm{s}^b$. 

\subsection{Global Sentence Encoding}
\label{s:gse}

As is known to us all, 
humans can leverage enormous prior knowledge to extract important parts for sentence semantics directly. 
However, it is quite formidable for models. They have to learn as much information as possible from the input data. 
Since RNN is powerful to process the sequence, and helps models to capture the intra-dependency and interaction of the input sequence, we select Stack-RNN~\cite{Nie2017ShortcutStackedSE} to process sentences. It is composed of multiple RNN layers on the top of each other. Note that we utilize Gated Recurrent Unit~(GRU) as the base unit. 
Specifically, let $\bm{H}_l$ be the $l^{th}$ GRU layer. At the time step $t$, Stack-RNN can be expressed as follows:
\begin{equation}
	\label{eq:ordinary-stack}
	\bm{h}_t^l = \bm{H}_l(\bm{x}_t^l, h_{t-1}^l), \quad \bm{x}_t^l = \bm{h}_t^{l-1},
\end{equation} 
where $\bm{x}_t^l$ is the input of the $t^{th}$ step in the $l^{th}$ GRU layer. While this architecture enables us to build up deeper representations, it cannot model sentence semantics comprehensively and preserve all the learned information, and even worse, this architecture might trigger explosion or make gradient problems vanish.  Thus, in order to optimize the capability of Stack-RNN and motivated by~\cite{Kim2018SemanticSM}, we concatenate the inputs $x^{(l-1)}$ and the states $h^{(l-1)}$ of the $(l-1)^{th}$ GRU layer as the inputs of the $l^{th}$ GRU layer and modified Eq.~(\ref{eq:ordinary-stack}) as follows:
\begin{equation}
	\label{eq:modified-stack}
	\bm{h}_t^l = \bm{H}_l(\bm{x}_t^l, h_{t-1}^l), \quad \bm{x}_t^l = [\bm{h}_t^{l-1}; \bm{x}_t^{l-1}],
\end{equation} 
where $[. ; .]$ denotes the concatenation operation. 
The final outputs are denoted as $\{\bm{h}_i^a|i=1, 2, ..., l_a\}$ and $\{\bm{h}_j^b|j=1, 2, ..., l_b\}$, which can preserve all the information, as well as the previous feature work in word embedding part. 

However, this architecture only models the sentence and stores all the information into vectors in a comprehensive way. How to compress these vectors into one sentence representation is still unclear. 
Since natural language has the redundancy mechanism~\cite{luuk2011redundancy}, different words have different contributions to sentence semantics. 
Moreover, self-attention can select the important parts at different positions from a single sequence~\cite{vaswani2017attention}.
Therefore, it is natural to leverage self-attention to generate sentence representations:
\begin{equation}
	\label{eq:sentence-attention}
	\begin{split}
		\bm{A}^a &= [\bm{h}_1^a, \bm{h}_2^a, ..., \bm{h}_{l_a}^a], \\
		\bm{\alpha}^a &= \bm{\omega}^{\mathrm{T}}\mathrm{tanh}(\bm{W}\bm{A}^a + \bm{b}), \\
		\bm{h}^a &= \sum_{i=1}^{l_a}\frac{\mathrm{exp}(\alpha_i^a)}{\sum_{k=1}^{l_a} \mathrm{exp}(\alpha_k^a)}\bm{h}_i^a, \quad i = 1, 2, ..., l_a,
	\end{split}
\end{equation} 
where $\{\bm{W}\in\mathbb{R}^{d_g * (d_w + l_s * d_g)}, \bm{\omega} \in \mathbb{R}^{d_g*1}\}$ are the trainable parameters. $d_g , d_w$ and $l_s$ denotes the hidden size of GRU, the embedding size of input words, and the stack layer number, respectively. 
$\bm{h}_i^a$ is the $i^{th}$ output of Stack-RNN for sentence $\bm{s}^a$. $\bm{h}^a$ denotes the global contextual representation of sentence $\bm{s}^a$, which is actually a weighted summation of the outputs of Stack-GRU. 
The same operation will be done on sentence $\bm{s}^b$ to get the global contextual representation $\bm{h}^b$.

\subsection{Dynamic Re-read Mechanism}
As mentioned in Section~\ref{s:introduction}, static usage of attention mechanism will inhibit its ability. 
Meanwhile, psychological research has shown that people usually pay close attention to the small part that they care about most, and dynamically change their focus point for an in-depth understanding of the sentence~\cite{wang1999reading}.  
Therefore, we develop the Dynamic Re-read~(DRr) Mechanism to solve the first challenge mentioned in Section~\ref{s:problem}. 
As shown in Fig.\ref{f:drr_model}(C), DRr unit selects the most important word at each step with the consideration of global contextual representations, and the selections in previous steps. 

In detail, the inputs of DRr unit are the final outputs $\{\bm{h}_i^a|i=1, 2, ..., l_a\}$ and $\{\bm{h}_i^b|i=1, 2, ..., l_b\}$ of previous component.  
In each reading step, we adopt attention mechanism to choose one important word $\bar{\bm{a}}_t$ from the whole input sequence based on the learned information (i.e., $\bar{\bm{h}}_{t-1}^a, \bm{h}^b$). Then, we utilize GRU to encode the chosen word and generate the dynamic re-read representation $\bm{v}^a$ for sentence $\bm{s}^a$ as follows:
\begin{equation}
	\label{eq:drr-basegru}
	\begin{split}
		\bar{\bm{a}}_t &= \mathrm{F}([\bm{h}_1^a, \bm{h}_2^a, ..., \bm{h}_{l_a}^a], \bar{\bm{h}}_{t-1}^a, \bm{h}^b), \\
		\bar{\bm{h}}_t^a &= \mathrm{GRU}({\bar{\bm{a}}_t, \bar{\bm{h}}_{t-1}^a}), \quad t = 1, 2, ..., T, \\
		\bm{v}^a &= \bar{\bm{h}}_T^a,
	\end{split}
\end{equation} 
where $\bm{h}^b$ is the global contextual representation for sentence $\bm{s}^b$. 
$T$ is the dynamic re-read length. 
We also employ the global contextual representation $\bm{h}^a$ as the initial state of GRU for sentence $a$. 
$\mathrm{F}(\cdot)$ is the choosing function, and we utilize attention mechanism to achieve this function as follows:
\begin{equation}
	\label{eq:dyn-unit}
	\begin{split}
		\bar{\bm{A}}^a &= [\bm{h}_1^a, \bm{h}_2^a, ..., \bm{h}_{l_a}^a],\\
		\bar{\bm{m}}^a &= \bm{\omega}_d^T\mathrm{tanh}(\bm{W}_d\bar{\bm{A}}^a + (\bm{U}_d\bar{\bm{h}}_{t-1}^a + \bm{M}_d\bm{h}^b)\otimes \bm{e}_{l_a}), \\
		\bar{\bm{\alpha}^a} &= \sum_{i=1}^{l_a}\frac{\mathrm{exp}(\bar{m}_i^a)}{\sum_{k=1}^{l_a} \mathrm{exp}(\bar{m}_k^a)}, \\
		\bar{\bm{a}}_t &= \bm{h}_j^a, \quad (j = \mathrm{Index}(\mathrm{max}(\bar{\bm{\alpha}^a}))), \\
	\end{split}
\end{equation} 
where $\{\bm{W}_d \in \mathbb{R}^{d_a * (d_w + l_s * d_g)}, \bm{U}_d \in \mathbb{R}^{d_a * d_g}, \bm{M}_d\in\mathbb{R}^{d_a * (d_w + l_s * d_g)}, \omega_d \in \mathbb{R}^{d_a*1}\}$ are trainable parameters. 
$ \bm{e}_{l_a} \in \mathbb{R}^{l_a} $ is a row vector of $ 1 $.
$\mathrm{Index}(\mathrm{max}(\bar{\bm{\alpha}^a}))$ denotes getting the corresponding index of the maximum value in the attention vector $\bar{\bm{\alpha}^a}$. 
The outer product $ (\bm{U}_d\bar{\bm{h}}_{t-1}^a + \bm{M}_d\bm{h}^b)\otimes \bm{e}_{l_a} $ means repeating $l_a$ times of the results $ (\bm{U}_d\bar{\bm{h}}_{t-1}^a + \bm{M}_d\bm{h}^b)$.  
Specifically, the global contextual representation $\bm{h}^b$ of sentence $\bm{s}^b$ is beneficial for choosing the most relevant information in sentence $\bm{s}^a$. 
The previous hidden state $\bar{\bm{h}}_{t-1}^a$ is capable of memorizing what has been selected in the previous steps. 
Treating them as the input of the attention unit, \drr~can select the most important part from sentence $\bm{s}^a$ at the $t^{th}$ time step with the consideration of the information from sentence $\bm{s}^b$ and the previous steps.  

However, $\mathrm{Index}(\mathrm{max}(\cdot))$ operation has no derivative, which means its gradient could not be calculated. Fortunately, our goal is to select the most important part, which requires one word at one step. Inspired by $\mathrm{softmax}$ function, we modify Eq.(\ref{eq:dyn-unit}) as follows to solve the non-derivative problem:
\begin{equation}
	\label{eq:modify-dyn-unit}
	\begin{split}
		\bar{\bm{A}}^a &= [\bm{h}_1^a, \bm{h}_2^a, ..., \bm{h}_{l_a}^a],\\
		\bar{\bm{m}}^a &= \omega_d^T\mathrm{tanh}(\bm{W}_d\bar{\bm{A}}^a + (\bm{U}_d\bar{\bm{h}}_{t-1}^a + \bm{M}_d\bm{h}^b)\otimes \bm{e}_{l_a}), \\
		\bar{\bm{a}}_t &= \sum_{i=1}^{l_a}\frac{\mathrm{exp}(\beta\bar{m}_i^a)}{\sum_{k=1}^{l_a} \mathrm{exp}(\beta\bar{m}_k^a)}\bm{h}_i^a,
	\end{split}
\end{equation} 
where $\beta$ is an arbitrarily big value. With this operation, the weight $\bar{\bm{a}}_t$ of the most important word will be very close to $1$, and other weights will be very close to $0$.

\subsection{Label Prediction}
This component consists of three operations: matching, fusion, and classification. 
In order to determine the overall relation between two sentences, we leverage heuristic matching~\cite{Chen-Qian2017ACL} between global contextual representations $\bm{h}^a$, $\bm{h}^b$ and dynamic re-read representations $\bm{v}^a$, $\bm{v}^b$. 
Specifically, we use the element-wise product, their difference and concatenation. 
Then, we send them to multi-layer perceptron (MLP) to calculate the relation probability between two sentences. 
The MLP has two hidden layers with $\mathrm{ReLU}$ activation and a $\mathrm{softmax}$ output layer.
\begin{equation}\label{eq:matching}
	\begin{split}
		\bm{h} &= [(\bm{h}^a;~\bm{h}^b);~(\bm{h}^b \odot \bm{h}^a);~(\bm{h}^b - \bm{h}^a)], \\
		\bm{v} &= [(\bm{v}^a;~\bm{v}^b);~(\bm{v}^b \odot \bm{v}^a);~(\bm{v}^b - \bm{v}^a)], \\
		\bm{p}^h &= \mathrm{MLP}_1(\bm{h}), \\
		\bm{p}^v &= \mathrm{MLP}_2(\bm{v}),
	\end{split}
\end{equation}
where $\bm{p}^h$ and $\bm{p}^v$ denote the probability distribution of different classes with global contextual sentence representations and dynamic important part representations, respectively.

After getting different probability distributions of the semantic relations using different sentence semantic representations, we intend to integrate this information to achieve more robust performance. 
Thus, we utilize a fusion gate and a multi-layer perceptron (MLP) to integrate $\bm{p}^h, \bm{p}^v$ and make the final classification, which can be formulated as follows:
\begin{equation}\label{eq:classify}
	\begin{split}
		\alpha_h &= \sigma(\bm{w}_h^\mathrm{T}\bm{p}^h + b_h), \\
		\alpha_v &= \sigma(\bm{w}_v^\mathrm{T}\bm{p}^v + b_v), \\
		P(y|(\bm{s}^a, \bm{s}^b)) &= \mathrm{MLP}_3(\alpha_h\bm{p}^h + \alpha_v\bm{p}^v).
	\end{split}
\end{equation}

\section{\fdlan}
\label{s:dlan}

As mentioned in Section~\ref{s:introduction}, our proposed \drr~still has some space for further improvement. 
First, the encoding capability of GRU has been proven weaker than transformer~\cite{devlin2018bert}. 
Replacing it with pre-trained models~(e.g., BERT) will import the incomplete modeling and fragile representation that BPE suffers from. 
Second, selecting only one important word at each reading time is insufficient since words with different contexts can express tremendously different meanings.

To solve the previous shortcomings, in this section, we focus on the local information utilization and extend \drr~to a novel \dlan, which employs pre-trained BERT to encode sentences and a newly designed Dynamic Sequential Attention~(DSA) for the usage of local structures and the selection of important parts.  
Our key contributions lie in two parts. 
The first is using a CNN-based structure to alleviate the problem that BPE suffers from and enhance the encoding ability of BERT.
The second is designing a novel DSA unit to integrate the chosen important part and corresponding local structure in a sequential way at each reading step for performance improvement of our proposed dynamic attention mechanism. 
Moreover, DSA unit utilizes coverage mechanism to ensure the diversity of dynamic selection, which can help to evaluate the semantic relation between the sentence pair more comprehensively. 
Next, we will introduce the technical details of \dlan.

\begin{figure*}
	\centering
	\includegraphics[width=0.8\textwidth]{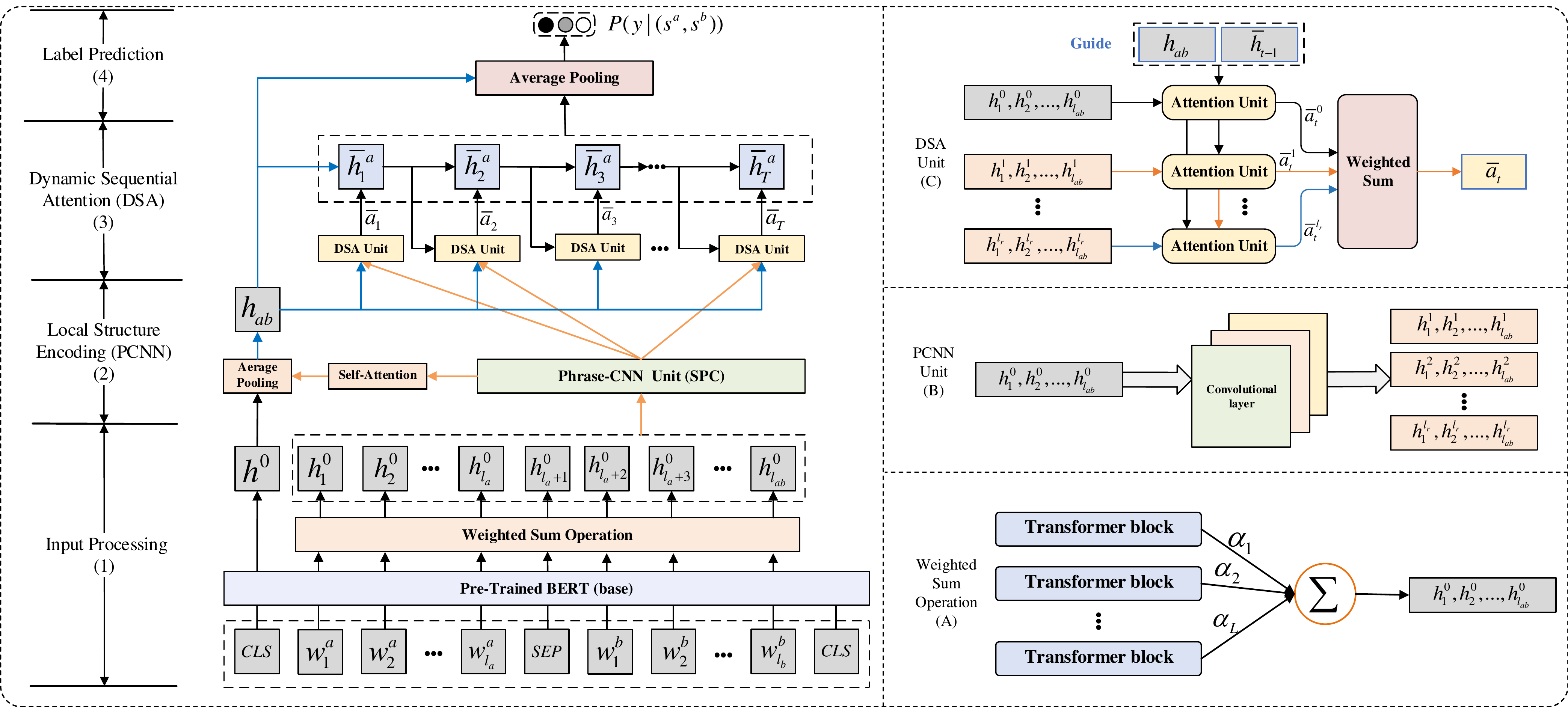}
	\caption{Overall Architecture of \fdlan~and Dynamic Sequential Attention.} 
	\label{f:dlan_model}
	\vspace{-4mm}
\end{figure*}

\subsection{Global Encoding}
\label{s:input}

In order to encoding sentences more comprehensively and make full use of pre-trained language model, \dlan~chooses BERT-base~\cite{devlin2018bert} to encode each word in the sentences from a global perspective. 
Specifically, the input sentence pair is first concatenated with \textit{``[SEP]''} token and added \textit{``[CLS]''} at the beginning and the end as BERT requires. 
Then, BPE tokenizer is employed to tokenize each word into BPE tokens. 
Suppose the final number of tokens in the sentence pair is $l_{ab}$, and BERT generates $L$ hidden states for each BPE token $\bm{BERT}_t^l, 1 \leq l \leq L, 1 \leq t \leq l_{ab}$. As shown in Fig.~\ref{f:dlan_model}(A), the contextual representation for $t^{th}$ token in input sentence pair at token level is then a per-layer weighted sum of transformer block output, with weights $\alpha_1, \alpha_2,...,\alpha_L$.
\begin{equation}
	\label{eq:input-processing}
	\begin{split}
		\bm{h}_t^0 &= \sum_{l=1}^{L}\alpha_l\bm{\bm{BERT}_t^l}, \quad 1 \leq t \leq l_{ab}, \\ 
	\end{split}
\end{equation} 
where $\alpha_l$ is the weight for the $l^{th}$ layer in BERT and is trained during model training. $\bm{h}^0_t$ is the contextual representation for the $t^{th}$ token. Here, $0$ denotes the index of weighted output from BERT and is consistent with the description in the following parts. Moreover, the output $\bm{h}^0$ of the first special token \textit{``[CLS]''} in the last block is treated as the contextual representation for input sentence pair at sentence level.

\subsection{Local Structure Encoding}
\label{s:tsse}
The semantic relation within the sentence pair is not only connected with the important words, but also affected by the corresponding local structure. 
Selecting only one important word at each reading step in \drr~seems insufficient for semantic relation modeling. 
Meanwhile, though BERT leverages multi-layer operation to perceive coarser granularity and learns more about word or phrase, the used BPE will import incomplete modeling and fragile representation problems for input sentences. 
To overcome the above shortcomings,  we adapt CNN with different composite kernels~(e.g., bigram and trigram) and propose the \emph{Phrase-CNN~(PCNN)} unit to extract local contexts of input sentence pair. 
As illustrated in Fig.~\ref{f:dlan_model}(B), let $\bm{h}_t^{r}$ be the local structure representation for the $t^{th}$ token with the $r^{th}$ kernel. 
This process can be formulated as follows: 
\begin{equation}
	\label{eq:phrase-cnn}
	\begin{split}
		\bm{h}_t^{r} = \mathrm{CNN}_r([\bm{h}^0_{t-k}, \bm{h}^0_{t-k+1},&...,\bm{h}^0_{t+k-1},\bm{h}^0_{t+k}]), \\ 
		&t = 1, 2,..., l_{r-1}, l_r, \\
	\end{split}
\end{equation} 
where $k$ denotes the kernel size of the $r^{th}$ \emph{PCNN} unit.
$l_r$ denotes the number of \emph{PCNN} units. 
In order to maintain consistency of the output of different encoding units in this layer, the output length~(i.e., number of output channels) of each \emph{PCNN} unit is the same as that of the input sequence. 
Moreover, we intend to generate the contextual representations $\bm{h}^r$ based on the local structure information from \textit{PCNN} more comprehensively. 
Thus, we apply self-attention to generate contextual representations over the output of \textit{PCNN}:
\begin{equation}
	\label{eq:dsa-sentence-attention}
	\begin{split}
		\bm{A}^{r}_{ab} &= [\bm{h}_1^{r}, \bm{h}_2^{r}, ..., \bm{h}_{l_{ab}}^{r}], \\
		\bm{\alpha}^{r} &=  \bm{\omega}_r^{\mathrm{T}}\mathrm{tanh}(\bm{W}_r\bm{A}^{r}_{ab} + b), \\
		\bm{h}^{r} &= \sum_{i=1}^{L}\frac{\mathrm{exp}(\alpha_i^{r})}{\sum_{j=1}^{L} \mathrm{exp}(\alpha_j^{r})}\bm{h}_i^{r}, \\  
		\bm{h}_{ab} &= \mathrm{avg}([\bm{h}^{0}, \bm{h}^1, \bm{h}^2,..., \bm{h}^{l_{ab}}]), \\
	\end{split}
\end{equation} 
where $\{\bm{W}_r\in\mathbb{R}^{d_a * 768},\bm{\omega}_r \in \mathbb{R}^{d_a * 1}\}$ are trainable parameters. 
$\bm{h}_i^{r}$ is the $i^{th}$ token representation from the $r^{th}$ \textit{PCNN} encoding kernel. 
$\mathrm{avg}$ denotes the average-pooling. 
$\bm{h}_{ab}$ denotes the sentence-level representation for input sentence pair.  
We have to note that we adopt BERT-base as the basic encoder. Thus, its output size $768$ for each token affects the parameter size of the following network structures of \dlan.

\subsection{Dynamic Sequential Attention~(DSA)}
In our preliminary work~\cite{zhang2019drr}, we leverage DRr unit to select one important word at each reading step. 
However, one word can express tremendously different semantic meanings due to different local structures. 
Only selecting one word may be insufficient for precise sentence semantic modeling. 
In order to leverage the corresponding local structures to enrich the analysis of the selected important parts and represent these parts more precisely, we develop a novel \textit{DSA} unit, which is shown in Fig.~\ref{f:dlan_model}(C).
\textit{DSA} employs the learned information to select the most important part in a sentence and uses the local structure to boost its representation in a sequential manner at each reading step.  
Then, a weighted sum operation is employed to integrate the selected important part and the corresponding local structure. 
After that, the enhanced result will be treated as the input of GRU.

Concretely, \textit{DSA} first employs the choosing function $\mathrm{F(\cdot)}$ and learned information~(i.e., $\bar{\bm{h}}_{t-1}, \bm{h}_{ab}, \bar{\bm{a}}_t^{[0:(k-1)]}$) to select the most important part $\bar{\bm{a}}_t^{0}$ from BERT output $[\bm{h}_1^{0}, \bm{h}_2^{0},...,\bm{h}_{l_{ab}}^{0}]$ at current step. 
Then, same operation is applied on \textit{PCNN} output ($[\bm{h}_1^{r}, \bm{h}_2^{r}, ..., \bm{h}_{l_{ab}}^{r}], r=1,2,...,l_r$) to select the proper local structures $\bar{\bm{a}}_t^{r}$ in a sequential manner, which can support the understanding to the important part. 
Next, it adopts a weight sum to fuse all these selected information.
Since humans generally select these important parts in a sequential manner, \emph{DSA} also selects GRU to process the fusion results. 
The fusion result $\bar{\bm{a}}_t$ is sent to a GRU and treated as the $t^{th}$ input of this GRU as follows: 
\begin{equation}
	\label{eq:dtsa-basegru}
	\begin{split}
		\bar{\bm{a}}_t^r &= \mathrm{F}([\bm{h}_1^{r}, \bm{h}_2^{r}, ..., \bm{h}_{l_{ab}}^{r}], \bar{\bm{h}}_{t-1}, \bm{h}_{ab}, \bar{\bm{a}}_t^{[0:(r-1)]}), \\
		\alpha_t^r &= \bm{\omega}_f^T \mathrm{tanh}(\bm{W}_f\bar{\bm{a}}_t^r), \\ 
		\bar{\bm{a}}_t &= \sum_{r=1}^{l_r}\frac{exp(\alpha_t^r)}{\sum_{j=1}^{l_r} exp(\alpha_t^j)}\bar{\bm{a}}_t^r, \\
		\bar{\bm{h}}_t &= \mathrm{GRU}({\bar{\bm{a}}_t, \bar{\bm{h}}_{t-1}}), \quad t = 1, 2, ..., T, \\
	\end{split}
\end{equation} 
where $\bar{\bm{a}}_t^{[0:(r-1)]}$ is previous $r$ selections at time step $t$. 
$\bar{\bm{h}}_{t-1}$ denotes the $(t-1)^{th}$ hidden state of GRU. 
$\bar{\bm{h}}_{0}$ is initialized with sentence-level representation $\bm{h}_{ab}$.
$T$ is the length of dynamic selection. 
$\mathrm{F(\cdot)}$ is the choosing function. 
We select the additive attention~\cite{Bahdanau2014NeuralMT} to implement $\mathrm{F(\cdot)}$ and modify it so that only the most important feature is selected:

\begin{equation}
	\label{eq:dlan-unit}
	\begin{split}
		\bar{\bm{A}}^{r,a} &= [\bm{h}_1^{r}, \bm{h}_2^{r}, ..., \bm{h}_{l_{ab}}^{r}], \\
		\bar{\bm{m}}_t^{r} &= \bm{\omega}_{rd}^T\mathrm{tanh}(\bm{W}_{rd}\bar{\bm{A}}^{r} + (\bm{U}_{rd}\bar{\bm{h}}_{t-1} \\ &+ \bm{M}_{rd}\bm{h}_{ab} + \sum_{j=0}^{r-1}\bm{V}_{rd}\bar{\bm{a}}_t^j)\otimes \bm{e}_{l_{ab}}), \\
		\bar{\bm{a}}_t^r &= \sum_{i=1}^{l_{ab}}\frac{\mathrm{exp}(\beta\bar{m}_{t,i}^{r})}{\sum_{j=1}^{l_{ab}} \mathrm{exp}(\beta\bar{m}_{t,j}^{r})}\bm{h}_i^{r},  \\
	\end{split}
\end{equation}
where $\{\bm{W}_{rd}\in\mathbb{R}^{d_a * 768}, \bm{U}_{rd} \in \mathbb{R}^{d_a * d_g}, \bm{M}_{rd}\in\mathbb{R}^{d_a * 768}, \bm{V}_{rd}\in\mathbb{R}^{d_a * 768}, \bm{\omega}_{rd}\in\mathbb{R}^{d_a * 1}\}$ are trainable parameters. 
Similar to \drr, we also modify the $\mathrm{softmax}$ with this arbitrary big value $\beta$ to imitate the process of the most important part selection. 
One step further, \textit{DSA} unit~selects the important parts of input sentences in an unsupervised manner, and this unit is learned along with the end-to-end training of entire model. 
Therefore, \textit{DSA} unit might result in two problems: 1) \textit{Over-selection}: some important parts are selected repetitively too many times; 2) \textit{Under-selection}: some important parts are mistakenly ignored.  
Inspired by coverage mechanism used in neural machine translation~\cite{tu2016modeling}, we also propose to leverage coverage mechanism to remember the dynamic selection at previous steps and it will be updated after each selection.  
Specifically, a coverage vector $\bm{c}_{t}^{r}$ is used to indicate the degree of which word is not concerned at the $t^{th}$ reading step for the $r^{th}$ encoding unit, enabling the diversity of dynamic selections in the model. 
This vector is initialized as an $l_{ab}$ dimension vector of $1$, indicating that all words in sentences have not been selected. 
After selecting the important part at each reading step, we update this vector based on the calculated attention weights $\bar{\bm{m}}_t^r$ and the hidden states $\bar{\bm{A}}^{r,a}$ of input sentences. 
Thus, Eq.~\ref{eq:dlan-unit} will be modified as follows:

\begin{equation}
	\label{eq:modified-dlan-unit}
	\begin{split}
		\bm{c}_{0}^{r} &= [1, 1, ..., 1] \in \mathbb{R}^{l_{ab}}, \\
		\bar{\bm{m}}_t^{r} &= \bm{c}_{t-1}^{r}\bm{\omega}_{rd}^T\mathrm{tanh}(\bm{W}_{rd}\bar{\bm{A}}^{r} + (\bm{U}_{rd}\bar{\bm{h}}_{t-1} \\ &+ \bm{M}_{rd}\bm{h}_{ab} + \sum_{j=0}^{r-1}\bm{V}_{rd}\bar{\bm{a}}_t^j)\otimes \bm{e}_{l_{ab}}), \\
		\bm{c}_t^{r} &= \bm{c}_{t-1}^{r} - \frac{1}{\phi}\bar{\bm{m}}_t^{r}, \quad
		\phi = T\cdot\sigma(\bm{W}_{\phi}\bar{\bm{A}}^{r}).
	\end{split}
\end{equation}

\subsection{Label Prediction}
\label{s:label-prediction}
After getting the dynamic reading results at each important position, it is natural to integrate the information for final prediction. 
Since the sentence-level output from BERT, PCNN, and hidden states from GRU indicate the semantic relations from different perspectives, we utilize average-pooling to process the results $[\bm{h}_{ab},\bar{\bm{h}}_1,\bar{\bm{h}}_2,...,\bar{\bm{h}}_T]$ for decision making. 
Then, we send the result $\bm{v}$ to a multi-layer perceptron~(MLP) for final classification. 
The MLP consists of one hidden layer with $\mathrm{ReLu}$ an activation and a $\mathrm{softmax}$ output layer.  
\begin{equation}
	\label{eq:fusion-classification}
	\begin{split}
		&\bm{v} = \mathrm{avg}([\bm{h}_{ab},\bar{\bm{h}}_1,\bar{\bm{h}}_2,...,\bar{\bm{h}}_T]), \\
		&P(y|(\bm{s}^a, \bm{s}^b)) = \mathrm{MLP}(\bm{v}). \\
	\end{split}
\end{equation}

\section{Experiment}
\label{s:experiment}
In this section, we first give a brief introduction about the datasets and evaluation methods. Then, we introduce implementation details about the proposed models. Next, we present empirical results on all datasets, and give a detailed analysis of the model and experimental results. For all reported results, we employ boldface and underline for the best and the second best results, respectively. 

\subsection{Datasets and Evaluation Methods}
To evaluate the proposed \drr~and \dlan~models, we conduct an empirical evaluation based on two well-known tasks (i.e., SNLI and PI). For each task, we select three benchmark datasets for evaluation. 
Specifically, for NLI task, we select SNLI~\cite{bowman2015large}, SICK~\cite{marelli2014semeval}, and SciTail~\cite{khot2018scitail} datasets to evaluate the model performance. For PI task, we choose Quora~\cite{iyer2017first}, MSRP~\cite{dolan2005automatically}, and Twitter-URL~\cite{lan2017continuously}. 
These two tasks cover the asymmetric and symmetric sentence relations separately, and can exhibit different characteristics.

Since both NLI and PI tasks can be treated as classification problems and most of baselines employ \textit{accuracy} as the evaluation metric, we also use \textit{accuracy} to compare the model performance. 
Moreover, we leverage \textit{F1} to testify the models on Twitter-URL dataset as other baselines did. 
We have to note that for each experiment, we repeat the evaluation process 5 times with different seeds, and report the best results.

\subsection{Implementation Details and Model Training}
In order to get the best performance, we leverage the validation set to tune the hyper-parameters.  
If the loss on validation set does not decrease in $1,000$ batches, we will stop the training process and accept the best model on validation as the final model.  
For MSRP~\cite{dolan2005automatically} and Twitter-URL~\cite{lan2017continuously} that do not have validation set, we randomly split $5\%$ data from training set as the validation set. 
Since the hyper-parameters will be various for different datasets, we list some common hyper-parameters of \drr~and \dlan~in Table~\ref{t:hyper-parameter}. 
They will be described in the following part.

For \drr, the word embedding size is set to $300$ and the character embedding size is set to $100$. 
The stack layer number in Global Sentence Encoding is $d_{s1}=3$. 
The Re-read length in dynamic re-read attention mechanism is $T_1=6$ . 
For initialization, we obtained the word embedding from a pre-trained word vectors (840B GloVe)~\cite{pennington2014glove} and the Out-Of-Vocabulary~(OOV) words are randomly initialized. They are kept to be fixed when training the entire model. 
All the model parameters are initialized with the uniform distribution in the range between $ -\sqrt{6/(nin+nout)} $ and $ \sqrt{6/(nin+nout)} $ as suggested by~\cite{orr2003neural}.  
All biases are set as zeros. 

For \dlan, the kernel sizes of \textit{PCNN} are set as $2$ and $3$, and the output dimensions of two-layer CNN are set as $500$ and $768$, the same as the output size of BERT. 
The length of dynamic selection in \textit{DSA}  is set as $5$. 
The optimizer we use is Adam optimizer and the initial learning rate is $0.001$. 

For both methods, we leverage Adam as the optimizer and the initial learning rate is $0.001$. 
During training, we design a learning rate decay methods called \textit{Cosine Warm Up Decay}, in which cosine operation and training epochs are utilized to update the learning rate in a non-linear manner. Corresponding details can be found in the released code\footnote{https://github.com/little1tow/Cosine-Warm-Up-Decay-method}.

\begin{table}
	\centering
	\caption{Hyper-parameters configuration in \drr~and \dlan.}
	\begin{footnotesize}
		\begin{tabular}{l|lr} \hline
			\textbf{Model} & \textbf{Hyper-parameters} & \textbf{Value} \\ \hline
			\multirow{6}{*}{\drr} 
			& Word embedding size & $d = 300$ \\
			& Character embedding size & $d_c = 100$ \\
			& embedding size of input words & $d_w = 412$ \\
			& GRU hidden size & $d_{g} = 256$ \\
			& attention size & $d_{a} = 200$ \\
			& Stack layer number & $l_{s} = 3$ \\
			& Dynamical Reading Length & $T_1 = 6$ \\
			& Initial learning rate & $\alpha = 0.001$ \\
			& Adam $\beta_1$ & $\beta_1 = 0.9$ \\ 
			& Adam $\beta_2$ & $\beta_2 = 0.999$ \\
			\hline 
			\multirow{4}{*}{\dlan} 
			& Kernel size of \textit{PCNN} & $d_k = 2, 3$ \\
			& Output size of \textit{PCNN} & $d_{c} = 500, 768$ \\
			& Dynamical Reading Length~& $T_2 = 5$ \\
			& Initial learning rate & $\alpha = 0.001$ \\
			\hline
		\end{tabular}
	\end{footnotesize}
	\label{t:hyper-parameter}
	\vspace{-4mm}
\end{table}

\begin{table*}
	\centering
	\caption{Experimental Results~(accuracy) on SNLI and Sick datasets.}
	\begin{tabular}{lccc} \hline
		\textbf{Model} & \textbf{SNLI Full test} & \textbf{SNLI Hard test} & \textbf{SICK test} \\ \hline
		(1) CAFE~\cite{Tay2017ACA} & 85.9\% & 66.1\% & 86.1\% \\
		(2) Distance-based SAN~\cite{im2017distance} & 86.3\% & 67.4\% & 86.7\% \\
		(3) DRCN~\cite{Kim2018SemanticSM} & 86.5\%& 68.3\% & 87.4\% \\ 
		(4) Dynamic Self-Attention~\cite{yoon2018dynamic} & 87.4\% & 71.5\%& 87.8\% \\ 
		(5) CSRAN~\cite{tay2018co} & 88.5\% & 76.8\% & 89.7\% \\
		(6) RE2~\cite{yang2019simple} & 88.9\% & 77.3\% & 89.8\% \\ \hline
		(7) BERT-(base)~\cite{devlin2018bert} & 90.3\% & 80.6\%& 88.7\% \\ 
		(8) BERT-(large)~\cite{devlin2018bert} & 90.7\% & 81.3\%& 88.3\% \\
		(9) RoBERTa-(base)~\cite{Liu2019RoBERTaAR} & 90.9\% & \underline{81.5\%} & \textbf{90.3\%} \\ 
		(10) ALBERT-(base)~\cite{Lan2020ALBERTAL} & 86.2\% & 77.5\%& 87.3\% \\ \hline
		(11) \drr & 87.7\% & 71.4\% & 88.3\% \\
		(12) \dlan-BERT-(base) & \underline{91.0\%} & 81.1\% & 89.5\% \\ 
		(13) \dlan-RoBERTa-(base) & \textbf{91.2\%} & \textbf{81.6\%} & \underline{89.8\%} \\ 
		\hline
	\end{tabular}
	\label{t:snli-result}
	\vspace{-4mm}
\end{table*}

\subsection{Natural Language Inference Results}

\subsubsection{Performance on SNLI and SICK}
\label{s:overall_results}
As illustrated in Table~\ref{t:snli-result}, for BERT-free methods, CSRAN~\cite{tay2018co} and RE2~\cite{yang2019simple} achieve the best performance. The following is our proposed \drr. 
Though \drr~did not perform the best, it leverage relatively simple structure~(DRr unit) to achieve competitive results, compared with the complex multiple RNN and attention calculation mixing layers used in CSRAN and RE2. 
This phenomenon supports the effectiveness of our proposed dynamic re-read attention. 
For powerful pre-trained models. 
We first note that \textit{``[CLS]''} output from the last layer of pre-trained models is leveraged for final classification, and we employ the same training strategy as our proposed methods did for a fair comparison. 
This is also the reason why results of pre-trained models are a little different from the GLUE leaderboard. 
From the table, we can first observe that BERT-based methods outperform all the BERT-free methods by a large margin, demonstrating the powerful encoding capability of pre-trained methods.  
Moreover, \dlan~performs better than these pre-trained models. 
Since they all leverage the pre-trained methods as the basic encoders, the performance improvement of \dlan~can demonstrate the usefulness of local structure. 
By employing the local structure, \dlan~can access the complete words or phrases in the input sentences, alleviating the incomplete modeling or fragile representation of BEP~\cite{Ma2020CharBERTCP} and achieving better performance. 
On the other hand, local structure and dynamic re-read attention utilization require more parameters to be trained. Thus, it may be a little underfitting when the dataset is small, which is the possible reason why \dlan~did not achieve the best performance on SICK dataset. The similar performance can also be observed on MSRP dataset from Table~\ref{t:pi-result}.

\begin{table}
	\centering
	\caption{Experimental Results~(accuracy) on SciTail dataset.}
	\begin{tabular}{lc} \hline
		\textbf{Model} & \textbf{SciTail test}\\ \hline
		(1) CAFE~\cite{Tay2017ACA} & 83.3\% \\
		(2) HBMP~\cite{talman2018natural} & 86.0\% \\ 
		(3) Finetuned Transformer LM~\cite{Radford2018ImprovingLU} & 88.3\% \\
		(4) CSRAN~\cite{tay2017compare} & 86.5\% \\
		(5) RE2~\cite{yang2019simple} & 86.2\% \\ \hline
		(6) BERT-(base)~\cite{devlin2018bert} & 93.1\% \\ 
		(7)  BERT-(large)~\cite{devlin2018bert} & 93.6\% \\
		(8)  RoBERTa-(base)~\cite{Liu2019RoBERTaAR} & \underline{93.8\%}  \\ 
		(9)  AlBERT-(base)~\cite{Lan2020ALBERTAL} & 91.4\% \\ \hline
		(10) \drr & 87.4\% \\
		(11) \dlan-BERT-(base) & \textbf{93.9\%} \\
		(12) \dlan-RoBERTa-(base) & \textbf{93.9\%} \\
		\hline
	\end{tabular}
	\label{t:scitail-result}
	\vspace{-4mm}
\end{table}

Among all baselines, CSRAN and RE2 are the current best BERT-free methods. RoBERTa-base~\cite{Liu2019RoBERTaAR} achieves the best performance among BERT based methods. 
After a detailed analysis, we conclude the following observations. First, both CSRAN and RE2 stack multi-layer blocks to achieve impressive performance, which is similar to the transformers in BERT based methods. 
This observation proves that stacking multi-layer blocks is indeed helpful for semantic representation. 
Second,  RoBERTa-base achieves better performance than BERT-large, which is counterfactual. 
We speculate the possible reason is that we did not fine-tune BERT-large well on these datasets, since there are huge amounts of parameters. 
Based on this reason, we only select BERT-base and RoBERTa-base as the basic encoders in \dlan.
Third, the improvement between results (13) and (9) is smaller than the improvement between results (12) and (7) in Table~\ref{t:snli-result}. 
This phenomenon supports the fact that these BERT-base models are not well trained~\cite{Liu2019RoBERTaAR}. 
Selecting RoBERTa-base as the basic encoder will be a better choice in sentence semantic modeling.

\subsubsection{Performance on SciTail Test}
Table~\ref{t:scitail-result} reports the experimental results on SciTail~\cite{khot2018scitail}. 
Compared with the other two NLI datasets,  SciTail dataset requires models to focus on entailment relation recognition. 
Moreover, both premise and hypothesis sentences are derived from the existing data source, which makes sentence pairs more linguistically challenging~\cite{khot2018scitail}. 
From the table, we can observe that even using multi-layer blocks, the performance of CSRAN is still $0.9\%$ lower than \drr. 
This observation indicates that our proposed dynamic re-read attention can deal with more complex and realistic situations. 
However, GRU used in \drr~still has some weaknesses in encoding input sentences, which leads \drr~to be $0.9\%$ lower than Finetuned Transformer~\cite{Radford2018ImprovingLU}. 
Compared with these BERT based methods, we can observe that \dlan~with BERT-base can outperform RoBERTa on this dataset, which not only indicates the usefulness of dynamic re-read attention, but also demonstrates that local structure can alleviate the weaknesses of BPE and boost the performance of dynamic re-read attention.

\begin{table}
	\centering
	\caption{Experimental Results~(accuracy) on Quora and MSRP datasets.}
	\begin{tabular}{lcc} \hline
		\textbf{Model} & \textbf{Quora test} & \textbf{MSRP test} \\ \hline
		(1) CENN~\cite{zhang2017context} & 80.7\% & 76.4\%\\
		(2) L.D.C~\cite{Wang2016SentenceSL} & 85.6\% & 78.4\%\\
		(3) BiMPM~\cite{Wang2017BilateralMM} & 88.2\% & -\\
		(4) DIIN~\cite{gong2017natural} & 89.1\% & -\\
		(5) DRCN~\cite{Kim2018SemanticSM} & 90.2\% & 82.5\%\\ 
		(6) RE2~\cite{yang2019simple} & 89.3\% & 78.5\% \\ \hline
		(7) BERT-(base)~\cite{devlin2018bert} & 91.0\% & 84.2\%\\ 
		(8) BERT-(large)~\cite{devlin2018bert} & \underline{91.4\%} & 85.4\%\\ 
		(9) RoBERTa-(base)~\cite{Liu2019RoBERTaAR} & 90.6\% & \underline{87.1\%} \\ 
		(10) ALBERT-(base)~\cite{Lan2020ALBERTAL} & 90.3\% & \textbf{88.6\% }\\ \hline
		(11) \drr & 89.8\% & 82.9\% \\
		(12) \dlan-BERT-(base) & \textbf{92.1\%} & 84.8\% \\ 
		(13) \dlan-RoBERTa-(base) & 91.3\% & 86.8\% \\
		\hline
	\end{tabular}
	\label{t:pi-result}
	\vspace{-4mm}
\end{table}

\subsection{Paraphrase Identification Results}
\label{s:pi-expeiment}
In addition to NLI task, we also select the PI task to verify the performance of our proposed models. Different from NLI task, PI task focus on the symmetric semantic relation and requires an agent to classifies whether two sentences express the same meaning, which is critical for question answering community, such as Quora and Zhihu\footnote{https://www.quora.com/, https://www.zhihu.com/}. 

\subsubsection{Performance on Quora and MSRP}
Tables~\ref{t:pi-result} reports the results on Quora and MSRP datasets. 
Still, our proposed methods achieve highly competitive performance by making full use of dynamic re-read attention and local structure. 
Besides, we observe that all methods have a better performance on Quora dataset than MSRP dataset.
One possible reason is that Quora dataset has more data~(over 400k sentence pairs) than MSRP dataset~(only $5,801$ sentence pairs).
In addition to the data size, we speculate that inter-sentence interaction is probably another possible reason. 
Quora dataset contains many sentence pairs with less complicated interactions~(e.g., many identical words in two sentences)~\cite{lan2018neural}. 
Thus, all the methods can achieve better performance on Quora dataset.
When analyzing the performance on MSRP, we can observe that ALBERT-base~\cite{Lan2020ALBERTAL} achieves the best performance on MSRP dataset. 
Meanwhile, our proposed \drr~outperforms RE2 by a large margin~($+4.4\%$). 
On the other hand, \dlan~with RoBERTa-base is not even as good as RoBERTa-base~($-0.3\%$). 
Since MSRP dataset only contains $5,801$ sentence pairs, relatively complex structures are harder to optimize well, let alone a better performance. 
This phenomenon is also consistent with the performance of our proposed models on SICK dataset, which is shown in Table~\ref{t:snli-result}.

\begin{table}
	\centering
	\caption{Experimental Results~(F1) on Twitter-URL dataset.}
	\begin{tabular}{lc} \hline
		\textbf{Model} & \textbf{Twitter-URL Test} \\ \hline
		(1) Shortcut-Stacked Encoder~\cite{Nie2017ShortcutStackedSE} & 0.650 \\
		(2) Decomposable Attention~\cite{Parikh2016ADA} & 0.652 \\
		(3) InferSent~\cite{conneau2017supervised} & 0.746 \\
		(4) $ESIM_{seq}$~\cite{Chen-Qian2017ACL} & 0.752 \\
		(5) PWIM~\cite{he2016pairwise} & 0.761 \\ 
		(6) RE2~\cite{yang2019simple} & 0.724 \\ \hline
		(7) BERT-(base)~\cite{devlin2018bert} & 0.765 \\ 
		(8) BERT-(large)~\cite{devlin2018bert} & 0.768 \\ 
		(9) RoBERTa-(base)~\cite{Liu2019RoBERTaAR} & 0.763 \\ 
		(10) ALBERT-(base)~\cite{Lan2020ALBERTAL} & 0.762 \\ \hline
		(11) \drr & 0.758 \\
		(12) \dlan-BERT-(base) & \textbf{0.772} \\ 
		(13) \dlan-RoBERTa-(base) & \underline{0.767} \\
		\hline
	\end{tabular}
	\label{t:twitter-result}
	\vspace{-4mm}
\end{table}

\subsubsection{Performance on Twitter-URL}
Table~\ref{t:twitter-result} shows the results (F1 value) on Twitter-URL dataset~\cite{lan2017continuously}. 
In addition to the impressive performance, we also find some interesting phenomena.
First of all, PWIM~\cite{he2016pairwise} is the best model among all BERT-free baselines, and even has a highly competitive performance compared with BERT based methods. 
By utilizing Bi-LSTM to encode the input sentences and a 19-layer deep CNN to integrate the features, PWIM makes a fully exploration on local structure on sentences and achieves a very competitive performance. 
However, the encoding and representing abilities of Bi-LSTM are still weaker than BERT. 
Thus, we observe that BERT based methods still outperform PWIM on this dataset. 
On the contrary, \dlan~not only selects BERT as the encoder to obtain comprehensive representations of input sentences, but also leverages local structure to alleviate the problems of BPE in BERT and strengthen the ability of our proposed dynamic re-read attention. 
Thus, we can observe that \dlan~achieves the best and the second best performances with different BERT-like encoders, surpassing these baselines.

\subsection{Ablation Performance}
The overall performance has proven the superiority of our proposed models. 
However, which part is more important for performance improvement is still unclear. 
Consequently, we conduct an ablation study on different test sets to examine the effectiveness of each component in \drr~and \dlan. 
Since it is very flexible to change different basic encoders in \dlan, we just examine the performance of \dlan~with BERT-base as encoder, and employ \dlan~to represent ``\dlan~with BERT-base'' for simplicity.

\subsubsection{Performance of \drr}
First of all, we select the global contextual sentence vector $\bm{h}^a$ as the initial vector of the GRU in dynamic re-read processing and $\bm{h}^b$ as the guide vector of dynamic selection operation. 
Therefore, \drr~can have a comprehensive understanding of sentence $\bm{s}^a$ and search the most relevant part with the consideration of sentence $\bm{s}^b$ and learned information at each step. 
When we remove each of them individually, we can obtain the results from Table~\ref{t:drr-ablation}~(1)-(2). 
These results demonstrate that the performance of both models decrease by nearly $2.0\%$, which is indicative of that initial vector and guide vector are critical for deciding which part should be concerned more at each step. 

Second, we are curious whether only one of the representations (i.e., global result or dynamic result) is enough for semantic matching. 
To this end, we remove the global results and re-read results from \drr. 
As illustrated in Table~\ref{t:drr-ablation}~(3)-(4), both results are extremely important for classification. 
Among them, global results play a more important role, which is consistent with human behavior. 
People generally read the text from the beginning to the end and then continue reading till reaching the end, so that they can understand the sentence semantics from a global perspective~\cite{zheng2019human}. 
Moreover, humans have the capability to leverage enormous prior knowledge to extract the important parts, which is quite arduous for models to access the knowledge. 
Therefore, the performance of input encoder is essential for our proposed models and a better encoder~(e.g., BERT) can boost the model performance.

\begin{table}
	\centering
	\caption{Ablation Performance (accuracy) of \drr.}
	\begin{footnotesize}
		\begin{tabular}{lcc} \hline
			\textbf{Model} & \textbf{SNLI test} & \textbf{SICK test} \\ \hline
			(1)~\drr~(w/o initial vector) & 85.3\% & 85.7\%  \\ 
			(2)~\drr~(w/o guide vector) & 85.1\% & 86.1\%  \\ \hline
			(3)~\drr~(w/o global result) & 81.2\% & 80.5\%  \\
			(4)~\drr~(w/o DRr result) & \underline{85.6\%} & \underline{86.4\%} \\ \hline
			(5)~\drr~ & \textbf{87.7}\%  & \textbf{88.3}\% \\ 
			\hline
		\end{tabular}
	\end{footnotesize}
	\label{t:drr-ablation}
	\vspace{-4mm}
\end{table}

\subsubsection{Performance of \dlan}
There are three main components in \dlan: 1) BERT encoder; 2) PCNN unit; 3) DSA unit. Therefore, we first examine their effectiveness by removing each of them. 
Table~\ref{t:dlan-ablation}~(1)-(4) report the corresponding results. 
We can observe that BERT encoder is still the most important component, which is consistent with the observation in the ablation performance in \drr.  
As a new scheme of NLP, BERT based methods can help models to access knowledge from large language corpora and surpass other NN methods by a large percentage. 

One of the main contributions of \dlan~is to model the local structure of input sentences. To verify the usefulness of local structure, we remove the PCNN unit from \dlan~and add the PCNN unit into BERT-base for the evaluation. 
Compared with the results between Table~\ref{t:dlan-ablation}~(1) and (2), we can observe that local structure can improve the performance of BERT-base. Moreover, by considering the local structure, BERT-base can achieve the second best performance on SNLI dataset. 
Meanwhile, the comparison between Table~\ref{t:dlan-ablation}~(3) and (7) also reports that the local structure can further boost the model performance. 
This phenomenon is also the same as the observation from Twitter-URL experiments in Section~\ref{s:pi-expeiment}. 
All of these indicate the usefulness of local structure. 
Further, when removing $\bm{h}_{ab}$ from the Label Prediction component and only taking the results of DSA unit into account, model performance has a big drop. 
Since DSA only focuses on the important parts and local contexts, it lacks global understanding of input sentences. 
Meanwhile, removing $\bm{h}_{ab}$ also means ignoring the superiority of BERT. 
Therefore, its performance would be greatly reduced.

\subsection{Sensitivity of Parameters}
\begin{figure*}
	\centering
	\includegraphics[width=0.9\textwidth]{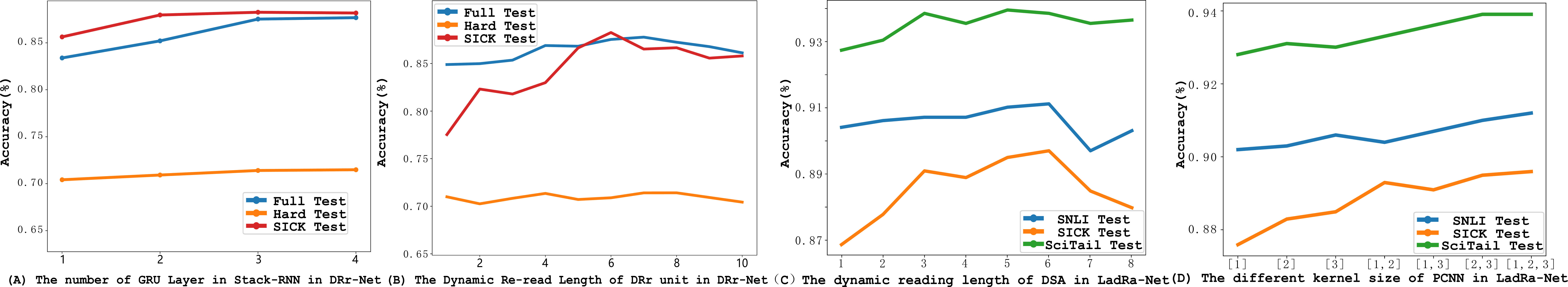}
	\caption{Performance~(accuracy) of \drr~and \dlan~with different hyper-parameters, including the number stack layers~(1-4) and reading lengths~(1-10) for \drr, and dynamical reading lengths~(1-8) and different kernel size (1-3) for \dlan. }
	\label{f:sensitive}
\end{figure*}

As mentioned in Sections~\ref{s:drr} and \ref{s:dlan}, two hyper-parameters affect the performance of each model separately: 
1) The stack layer number $l_{s}$ of Stack-RNN and the dynamic reading length $T_1$ of DRr for \drr; 
2) The kernel size $d_k$ of PCNN and the dynamic reading length $T_2$ of DSA for \dlan.
Therefore, we evaluate their impact on NLI task with different hyper-parameter settings. 
The results are reported in Fig.\ref{f:sensitive}. 

First of all, the most important hyper-parameter that both models have is the dynamic reading length~(i.e., $T_1, T_2$). 
We observe from Fig.\ref{f:sensitive}(B) that the performance of \drr~first becomes better with the increasing of re-read length. 
When the re-read length $T_1$ is between $5$ to $7$, \drr~achieves the best performance. 
This phenomenon is consistent with the psychological findings that human attention focuses on nearly 7 words~\cite{tononi2008consciousness}. 
When the length is bigger than $7$, the accuracy of \drr~decreases to varying degrees.
In other words, too short a reading length may be possible to cause the model to ignore some important parts, and too long a reading length may weaken the capability of capturing the important part, as well as understanding the semantic meaning intensively.

When it comes to \dlan, we can obtain similar phenomenon from Fig.\ref{f:sensitive}(C). 
Different from \drr, the best shot is around $T_2 = 5$, and \dlan~is more sensitive to the dynamic reading length. 
With the help of BERT based encoder and local structure utilization, \dlan~can achieve better performance with short re-read length and save some time. 
However, \dlan~is more complex than \drr, especially the BERT-base encoder. 
We need to tune the hyper-parameters carefully for the model performance.  

Second, Fig.\ref{f:sensitive}(A) reports the model performance with different $l_{s}$. 
We can observe that the performance becomes better with the increasing of $l_{s}$. 
With the continuous increasing of GRU layers, the increasing rate of accuracy will slow down, and even become worse on some test sets. 
With the consideration of model architecture, the scale of parameters will grow rapidly with the increasing of GRU layers, which may trigger the model hard to optimize, and even worse the gradient might explode or vanish. 
Moreover, the encoding capability of GRU is not comparable with BERT. To this end, we replace GRU with BERT-base in \dlan.

Third, Fig.\ref{f:sensitive}(D) illustrates the performance of \dlan~with different combinations of kernel size $d_k$.  
We observe that the model performance will increase with the increasing of $d_k$. 
As mentioned before, \textit{PCNN} is capable of capturing the local structure to enhance the semantic modeling.  
This result demonstrates that a large kernel size can help \dlan~to consider more local information and also proves the usefulness of the local structure. 
However, the effect of $d_k=1$ is not obvious. 
With the consideration of model complexity, we select $2$ and $3$ as our final kernel sizes. 

\begin{table}
	\centering
	\caption{Ablation performance (accuracy) of \dlan.}
	\begin{tabular}{lcc} \hline
		\textbf{Model} & \textbf{SNLI test} & \textbf{SciTail test} \\ \hline
		(1)~BERT-base & 90.3\% & 93.1\% \\
		(2)~BERT + PCNN & \underline{90.6\%} & 93.3\% \\
		(3)~BERT + DSA & 90.4\% & 92.7\% \\ 
		(4)~Without $\bm{h}_{ab}$ & 85.6\% & 82.7\%\\ \hline
		(5)~Stack-GRU + PCNN + DSA & 88.2\% & 87.9\%\\ 
		(6)~Max-pooing fusion & \textbf{91.0\%} & \underline{93.7\%} \\ \hline
		(7)~\dlan~(BERT+ PCNN + DSA) & \textbf{91.0}\% & \textbf{93.9}\%\\ 
		\hline
	\end{tabular}
	\label{t:dlan-ablation}
	\vspace{-4mm}
\end{table}

\begin{table*}
	\centering
	\caption{Some examples of re-read sequence and the classification.}
	\begin{footnotesize}
		\begin{tabular}{p{7.1cm}|p{5.7cm}|c|c} \hline
			\textbf{Sentence} & \textbf{Re-read sequence} & \textbf{Gold Label} & \textbf{Predicted Label} \\ \hline
			$\bm{s}^a$: a couple walk hand in hand down a street. & walk walk couple couple street street & \multirow{2}{*}{Contradiction} & \multirow{2}{*}{Contradiction} \\ \cline{1-2}
			$\bm{s}^b$: a couple is sitting on a bench. & couple sitting sitting bench bench bench \multirow{2}{*}{} & \multirow{2}{*}{} \\ \hline
			
			$\bm{s}^a$: a person in a red shirt and black pants hunched over. & red shirt red shirt red shirt & \multirow{2}{*}{Entailment} & \multirow{2}{*}{Entailment} \\ \cline{1-2}
			$\bm{s}^b$: a person wears a red shirt. & wears red shirt red shirt red \multirow{2}{*}{} & \multirow{2}{*}{} \\ \hline
			
			$\bm{s}^a$: a person with a purple shirt is painting an image of a woman on a white wall. & person painting painting woman woman woman & \multirow{2}{*}{Neutral} & \multirow{2}{*}{Neutral} \\ \cline{1-2}
			$\bm{s}^b$: a woman paints a portrait of her best friend. & woman paints paints best best best \multirow{2}{*}{} & \multirow{2}{*}{} \\ \hline \hline
			
			$\bm{s}^a$: the man in the black t-shirt is trying to throw something. & black t-shirt is trying throw throw  & \multirow{2}{*}{Entailment} & \multirow{2}{*}{Neutral} \\ \cline{1-2}
			$\bm{s}^b$: the man is in a black shirt. & black black shirt shirt shirt shirt \multirow{2}{*}{} & \multirow{2}{*}{} \\ 
			\hline
			
			$\bm{s}^a$: a person with a purple shirt is painting an image of a woman on a white wall. & person painting painting woman woman woman  & \multirow{2}{*}{Neutral} & \multirow{2}{*}{Entailment} \\ \cline{1-2}
			$\bm{s}^b$: a woman paints a portrait of a person. & paints portrait portrait portrait portrait person & \multirow{2}{*}{} & \multirow{2}{*}{} \\ 
			\hline
		\end{tabular}
	\end{footnotesize}
	\label{t:test-examples}
	\vspace{-4mm}
\end{table*}

\subsection{Case Study and Error Analysis}
To visually demonstrate the validity of \drr~and \dlan, we give some random examples from SNLI dataset. 
Since the core idea of \drr~and \dlan~is the Dynamic Re-read attention. For simplicity, we list the dynamic reading sequences of \drr~on these examples. The results are shown in Table~\ref{t:test-examples}. 

For the first example in Table~\ref{t:test-examples}, \drr~pours attention to words ``\textit{walk, couple, street}'' in sentence $\bm{s}^a$ and ``\textit{couple, sitting, bench}'' in sentence $\bm{s}^b$. 
Then, \drr~repeatedly processes these important words for the final decision. 
From these words, we can conclude that the relation of this sentence pair was contradiction easily. 
Moreover, when checking the entailment relation in the second example, \drr~processes the same important words repeatedly, i.e., reading ``red shirt'' multiple times. 
In other words, \drr~does choose the important region and re-read these important parts multiple times for the final decision. 
In order to better verify the ability of \drr, we make a error analysis on the misclassification examples. 
In the fourth example, when sentence $\bm{s}^a$ contains more information than sentence $\bm{s}^b$, \drr~may consider the unseen information more and make a misclassified decision. 
Moreover, when the sentence pair has very complex semantic relations, e.g., the last example in Table~\ref{t:test-examples}, the model may be confused about their semantics and suffer from one of the important words, which leads to a wrong classification result. 

\section{Conclusion and Future Work}
\label{s:conclusion}
In this paper, we presented a study on sentence semantic representation and matching. 
Specifically, we investigated that most attention-based methods attempted to select all the important parts in a static way, restricting the ability of attention mechanism. 
Inspired by the dynamic embedding methods used in pre-trained language models and the observation that people would constantly change their focal point for in-depth understanding of sentences, we proposed a newly designed dynamic re-read attention, which is able to pay close attention to a small region of sentences at each time and re-read the important parts for better sentence semantic matching. 
Based on this attention mechanism, we developed a novel \drr~for sentence semantic matching. 
Moreover, GRU encoder used in \drr~still has some weaknesses in representing sentence semantics. 
Therefore, we employ a pre-trained language model~(e.g., BERT, RoBERTa) as the input encoder. 
One step further, the BPE used in pre-trained language models still suffered from incomplete modeling and fragile representation problems. 
In order to alleviate these problems and further boost the ability of dynamic re-read attention, we proposed to take local structures of sentences into consideration and extended \drr~to \dlan. 
Finally, extensive experiments on two sentence semantic matching tasks (i.e., NLI and PI) demonstrated that \drr~could significantly improve the performance of sentence semantic matching. 
Furthermore, \dlan~was able to model sentence semantics more precisely and achieve better performance by considering the local structures of sentences. 

In the future, we will focus on providing more information for dynamic re-reading attention to better sentence semantic representing and matching. 
We also hope that our work could inspire the relative research and lead to many future works. 

\section*{Acknowledgment}

This research was partially supported by grants from the Young Scientists Fund of the National Natural Science Foundation of China (No. 62006066), the National Natural Science Foundation of China (No. 61727809, 61922073, and 61972125), the Fundamental Research Funds for the Central Universities, HFUT (JZ2021HGTB0075 and JZ2020HGPA0114) the Open Project Program of the National Laboratory of Pattern Recognition (NLPR).

\ifCLASSOPTIONcaptionsoff
  \newpage
  
\fi



%
\bibliographystyle{IEEEtran}
\bibliography{10_myReference}

\vspace{-14mm}
\begin{IEEEbiography}[{\includegraphics[width=1in,height=1.25in,clip,keepaspectratio]{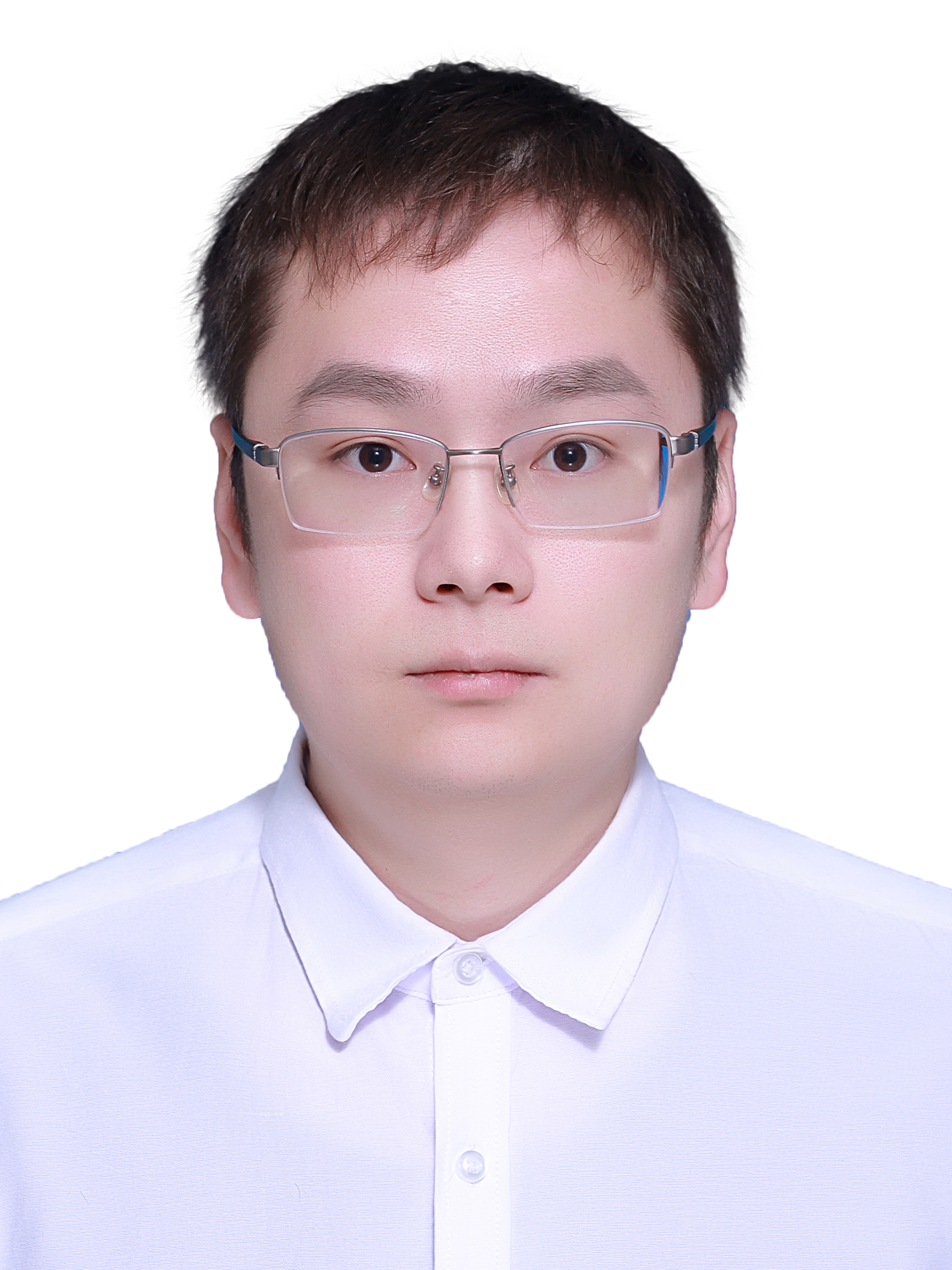}}]{Kun Zhang}
	received the PhD degree in computer science and technology from University of Science and Technology of China, Hefei, China, in 2019. He is is currently a faculty member with the Hefei University of Technology (HFUT), China. His research interests include Natural Language Understanding, Recommendation System. He has published several papers in refereed journals and conferences, such as the IEEE TSMC:S, IEEE TKDE, ACM TKDD, AAAI, KDD, ACL, ICDM. He received the KDD 2018 Best Student Paper Award.
\end{IEEEbiography}
\vspace{-16 mm}
\begin{IEEEbiography}[{\includegraphics[width=1in,height=1.25in,clip,keepaspectratio]{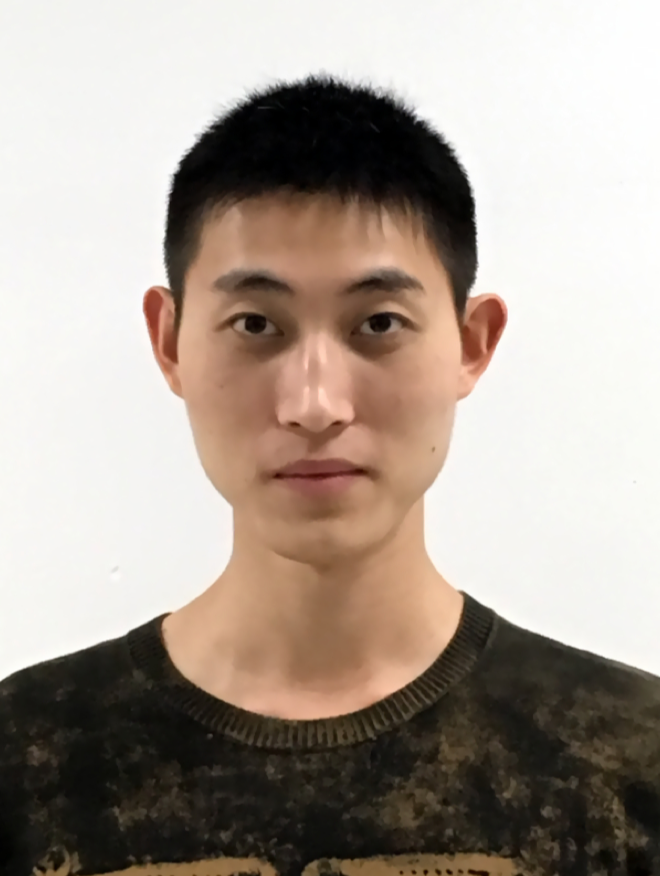}}]{Guangyi Lv}
	received the Ph.D. degree in computer science and technology from University of Science and Technology of China, Hefei, China, in 2019. He is currently an Advirosy Researcher with AI Lab at Lenovo Research, Beijing, China. His major research interests include natural language processing, computer vision and lifelong learning. He has published over 20 papers in refereed conference proceddings, such as AAAI, IJCAI, ICDM, DASFAA and PAKDD. 
\end{IEEEbiography}
\vspace{-15mm}
\begin{IEEEbiography}[{\includegraphics[width=1in,height=1.25in,clip,keepaspectratio]{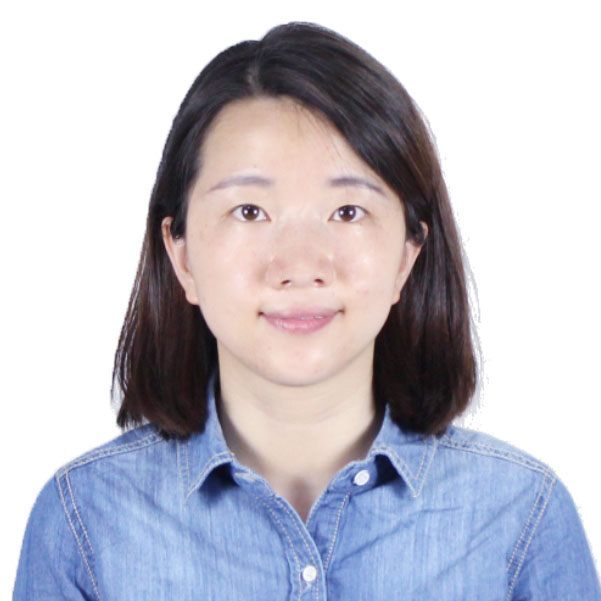}}]{Le Wu}
	is currently an associate professor and Ph.D. supervisor at the Hefei University of Technology (HFUT), China. She received the Ph.D. degree from the University of Science and Technology of China (USTC). Her general area of research interests is data mining, recommender systems and social network analysis. She has published more than 50 papers in referred journals and conferences. Dr. Le Wu is the recipient of the Best of SDM 2015 Award, the Distinguished Dissertation Award from China Association for Artificial Intelligence (CAAI) 2017, and the Youth Talent Promotion Project from China Association for Science and Technology.
\end{IEEEbiography}
\vspace{-18mm}
\begin{IEEEbiography}[{\includegraphics[width=1in,height=1.25in,clip,keepaspectratio]{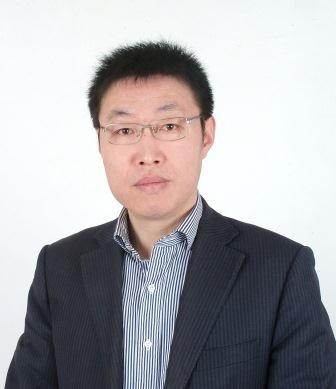}}]{Enhong Chen}
	(SM’07) is a professor and vice dean of the School of Computer Science at University of Science and Technology of China (USTC). He received the Ph.D. degree from USTC. His general area of research includes data mining and machine learning, social network analysis and recommender systems. He has published more than 100 papers in refereed conferences and journals, including IEEE TKDE, IEEE TMC, KDD, ICDM, NIPS, and CIKM. He was on program committees of numerous conferences including KDD, ICDM, SDM. He received the Best Application Paper Award on KDD-2008, the Best Student Paper Award on KDD-2018 (Research), the Best Research Paper Award on ICDM2011 and Best of SDM-2015. His research is supported by the National Science Foundation for Distinguished Young Scholars of China. He is a senior member of the IEEE.
\end{IEEEbiography}
\vspace{-18mm}
\begin{IEEEbiography}[{\includegraphics[width=1in,height=1.25in,clip,keepaspectratio]{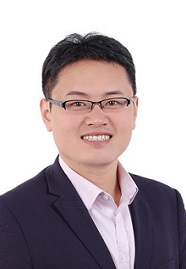}}]{Qi Liu}
	is an associate professor at University of Science and Technology of China (USTC). He received the Ph.D. degree in Computer Science from USTC. His general area of research is data mining and knowledge discovery. He has published prolifically in refereed journals and conference proceedings, e.g., TKDE, TOIS, TKDD, TIST, KDD, IJCAI, AAAI, ICDM, SDM and CIKM. He has served regularly in the program committees of a number of conferences, and is a reviewer for the leading academic journals in his fields. He is a member of ACM and IEEE. Dr. Liu is the recipient of the KDD 2018 Best Student Paper Award (Research) and the ICDM 2011 Best Research Paper Award. He is supported by the Young Elite Scientist Sponsorship Program of CAST and the Youth Innovation Promotion Association of CAS.
\end{IEEEbiography}
\vspace{-150mm}
\begin{IEEEbiography}[{\includegraphics[width=1in,height=1.25in,clip,keepaspectratio]{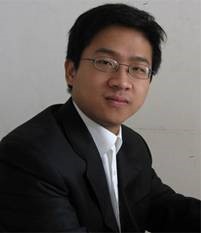}}]{Meng Wang}
	(F'21) received the BE and PhD degrees from USTC, in 2003 and 2008, respectively. He is a professor with HFUT. His current research interests include multimedia content analysis, computer vision, and pattern recognition. He has authored more than 200 book chapters, journal, and conference papers in these areas. He is the recipient of the ACM SIGMM Rising Star Award 2014. He is an associate editor of the IEEE Transactions on Knowledge and Data Engineering, the IEEE Transactions on Circuits and Systems for Video Technology, and the IEEE Transactions on Neural Networks and Learning Systems. He is an IEEE Fellow and IAPR Fellow.
\end{IEEEbiography}




\end{document}